\documentclass[final,5p,times,twocolumn]{elsarticle}

\usepackage{amssymb}

\usepackage{graphicx}

\usepackage{url}
\usepackage{times}
\usepackage{epsfig}
\usepackage{amsmath}
\usepackage{amssymb}
\usepackage{comment}

\usepackage{subcaption}
\usepackage{multirow}
\usepackage[table]{xcolor}
\usepackage{booktabs}
\usepackage{stmaryrd}

\journal{Image and Vision Computing}

\begin{document}

\begin{frontmatter}



\title{Unsupervised Domain Adaptation for Mobile Semantic Segmentation based on Cycle Consistency and Feature Alignment}


\author{Marco Toldo}
\author{Umberto Michieli}
\author{Gianluca Agresti}
\author{Pietro Zanuttigh}

\address{Department of Information Engineering, University of Padova, 35131, Padova, Italy}

\begin{abstract}
The supervised training of deep networks for semantic segmentation requires a huge amount of labeled real world data. To solve this issue, a commonly exploited workaround is to use synthetic data for training, but deep networks show a critical performance drop when analyzing data with slightly different statistical properties with respect to the training set.  In this work, we propose a novel Unsupervised Domain Adaptation (UDA) strategy to address the domain shift issue between real world and synthetic representations. An adversarial model, based on the cycle consistency framework, performs the mapping between the synthetic and real domain. The data is then fed to a MobileNet-v2 architecture that performs the semantic segmentation task. An additional couple of discriminators, working at the feature level of the MobileNet-v2, allows to better align the features of the two domain distributions and to further improve the performance. Finally, the consistency of the semantic maps is exploited. After an initial supervised training on synthetic data, the whole UDA architecture is trained end-to-end considering all its components at once. Experimental results show how the  proposed strategy is able to obtain impressive performance in adapting a segmentation network trained on synthetic data to real world scenarios. The usage of the lightweight MobileNet-v2 architecture allows its deployment on devices with limited computational resources as the ones employed in autonomous vehicles.
\end{abstract}

\begin{keyword}
Unsupervised Domain Adaptation \sep 
Semantic Segmentation \sep
Adversarial Learning \sep
Transfer Learning \sep
Image-to-Image Translation



\end{keyword}

\end{frontmatter}

\section{Introduction}
\label{sec:intro}

Semantic segmentation is one of the most challenging tasks in the computer vision field. All the current state-of-the-art approaches exploit deep learning architectures and  are typically based on fully convolutional models with an encoder-decoder architecture.
Deep neural networks allow to obtain impressive performance, however they require a huge amount of labeled data for their training. This is even more critical in semantic segmentation, where a pixel-wise labeling of the images is required and this operation is very expensive and time-consuming to be accomplished on a large scale.

A recently introduced workaround \cite{Richter2016,ros2016} is to use computer generated data for training,
but, despite the high level of realism of recent graphic engines, there is still a large domain shift between computer generated images and real world scenes. 
This domain shift issue needs to be tackled to obtain reliable performance in real world settings. Many different approaches for this task have been presented and it is possible to group them into two main families.

The methods belonging to the first family are based on the idea of performing the adaptation at image level before feeding the images to the network. Thus, a sort of style transfer operation is used to map each image from the source domain to a corresponding representation in the target domain. Most of these approaches are based on a generative adversarial model, in particular variants of the Cycle-GAN approach \cite{cyclegan}.

The methods belonging to the second family, instead, try to adapt the features produced by the network on the target domain in an unsupervised way mostly using adversarial learning techniques.  This adaptation can be performed both at the output level \cite{tsai2018, agresti2019unsupervised} or can involve internal network representations \cite{hoffman2016}.

In this work, we introduce an efficient Unsupervised Domain Adaptation (UDA) approach combining both strategies: we start performing an image-level domain adaptation using a model based on the Cycle-GAN framework that converts the input synthetic images to the  target (real) domain while preserving the semantic content. Then, the data is sent to a MobileNet-v2 architecture \cite{mobilenetv2} that performs the semantic segmentation.  In order to enhance the feature-level adaptation, an additional couple of discriminators working at the intermediate feature level of the network is employed.  Moreover, an additional loss component forces also the consistency between the semantic maps, thus avoiding the risk that the domain translation affects the semantic content.
Differently from other competing approaches that train independently the various sub-components, we train the complete architecture end-to-end on both synthetic labeled data  and  unlabeled real world data  in a single optimization framework based on adversarial learning.
Finally, using the simple and fast MobileNet-v2 architecture, the inference stage of the approach is suitable for real-time applications as the autonomous driving scenario chosen for the experimental evaluation.

The experimental evaluation, performed using the synthetic datasets SYNTHIA \cite{ros2016} and GTA5 \cite{Richter2016} and the real world dataset Cityscapes \cite{Cordts2016}, shows how the proposed framework is able to achieve large performance gains on real world datasets without using any labeled real world data during training. 

The reminder of the paper is organized as follows: Section \ref{sec:related} presents the related works. The proposed method is described in detail in Section \ref{sec:method}, while the training procedures are presented in Section \ref{sec:training}. Finally, the experimental evaluation is discussed in Section \ref{sec:results}, and Section \ref{sec:conclusions} draws the conclusions.

\section{Related Work}
\label{sec:related}
Unsupervised domain adaptation in deep learning has been traditionally addressed by exploiting a measure of domain dissimilarity between source and target distributions. Some works refer to the Maximum Mean Discrepancy (MMD), such as \cite{TzengHZSD14, long2015learning, LongZ0J16}. Tzeng et al.\ \cite{TzengHZSD14} introduce an adaptation layer and a domain confusion loss in the standard CNN architecture to learn domain invariant representations. In \cite{long2015learning,LongZ0J16} the degradation in transferability of application-specific hidden representations is tackled by matching mean domain embeddings in a new space. A different approach relies on correlation alignment, taking into account second order statistic  properties of the dataset \cite{SunFS16,SunS16}. A key factor for the majority of these works is that the deep network adaptation is performed end-to-end by assisting the task loss with a supplementary adaptation objective. Furthermore, no restriction to a specific network architecture is assumed. 

\textbf{Domain adaptation with adversarial learning.}\hspace{0.8mm} A game-changer has been the introduction of adversarial learning to perform domain adaptation. 
The key idea is to bridge the domain gap between source and target distributions by means of a domain classifier, where the discriminative action is applied on data samples (or their feature representations) from different domains. 
The adversarial strategy has been applied directly inside the input space (pixel-level) or on an intermediate latent space (feature-level). In the first case, the generator performs cross-domain mapping of images from the source to the target domain in order to generate a form of target supervision. When dealing with feature-level adaptation, the adversarial objective instead seeks a distribution alignment of source and target latent representations. 
Several works have explored unsupervised domain adaptation with a generative perspective \cite{cyclegan, LiuT16, bousmalis2017, ShrivastavaPTSW17}. Some of them resort to an image-to-image translation network, such as the CycleGAN \cite{cyclegan} and CoGAN \cite{LiuT16} architectures. 
Other works focus specifically on synthetic data adaptation \cite{bousmalis2017, ShrivastavaPTSW17}. 
A different line of research has been focusing on feature-space adaptation \cite{ganin2015, ganin2016, tzeng2017, ren2018}. 
One of the first contributions is brought by Ganin et al.\ \cite{ganin2015, ganin2016}. They propose a solution based on a gradient reversal layer to effectively back-propagate the error information during the adversarial min-max game. 
Following a similar approach, Tzeng et al.\ \cite{tzeng2017} introduce the ADDA model, where the feature extractor is no longer shared by the source and target domains. 
In order to learn more transferable representations, Ren et al.\ \cite{ren2018} explore adversarial feature adaptation in the context of multi-task learning. 

\textbf{UDA for Semantic Segmentation.}\hspace{0.8mm}
The majority of the aforementioned works, despite being devised as application-independent, are mainly focused on the object detection and image classification tasks. 
Unlike those tasks, semantic segmentation involves complex high-dimensional representations to capture spatial affinity of local semantics. In addition, semantic segmentation predictions belong to a very large label space due to their dense pixel-level highly-structured nature. For these reasons, domain adaptation in the context of semantic segmentation requires an extra effort in model design, as well as supplementary techniques to deal with the inherent additional complexity.
Hoffman et al.\ \cite{hoffman2016} have been the first to propose an adaptive framework for semantic segmentation. Similarly to previous methods 
devised for image classification, an unsupervised adversarial approach is employed to perform global domain alignment at feature level. 
A local alignment constraint-based technique is also introduced to acquire category specific knowledge from the target domain and achieve a more robust adaptation. 
A different path is followed by Zhang et al.\ \cite{zhang2017}, who adopt a curriculum-style adaptive method. First, some high level label properties are inferred on target data and then the semantic segmentation hard task is learned with a standard source-based supervised optimization, assisted by additional losses to enforce the inferred properties on target predictions lacking of supervision.
Curriculum domain adaptation is further explored in combination with self-training in \cite{lian2019constructing}.\\
Recently, 
many researches have tried to improve the state-of-the-art techniques for domain adaptation in the semantic segmentation field, especially focusing on adaptation from synthetic data to real world scenes. Some of those contributions resort to style transfer algorithms to artificially generate target supervision \cite{Dundar18, Wu19}.
Another line of work falls in the category of adversarial adaptation methods \cite{hoffman2018, sankaranarayanan2018, Sankaranarayanan18a, tsai2018, MurezKKRK18, VuJBCP19, ChenL0H19, pizzati2019domain}. 
Sankaranarayanan et al.\ \cite{sankaranarayanan2018, Sankaranarayanan18a} use a generative adversarial module to learn an encoder-decoder couple capable of projecting images into a feature space and reconstructing them back for both source and target domains. At the same time, the entire network is trained to produce target-like images when features are extracted from input source samples. 
Tsai et al.\ \cite{tsai2018} investigate adversarial learning in the output space (i.e., segmentation maps) to learn domain invariant semantic representations. They also extend their model with a multi-level adaptation framework to adapt features at different scales. 
Similarly, Murez et al.\ \cite{MurezKKRK18} design an adaptation framework based on a backbone encoder supported by multiple auxiliary networks and losses with regularization purposes to achieve domain agnostic feature extraction. The discovered latent feature space is then exploited to learn a domain invariant predictor. 
Feature alignment of the two domains in the latent feature space is further developed in \cite{luo2019significance} through an information bottleneck before the adversarial adaptation module on the feature space. 
Following a self-training strategy, Vu et al.\ \cite{VuJBCP19} resort to an entropy-based loss. 
They explore both a direct approach with a hand-crafted loss and indirect adversarial solution for which the objective is expressed by a learnable discriminator. Other works \cite{michieli2019adversarial, biasetton2019, HungTLL018} resort to a self-training technique by generating pseudo-labels from high-scoring pixel-wise predictions selected by applying a threshold on probability maps computed on unlabeled data. 
In \cite{vu2019dada} an additional task module using dense depth information is exploited to boost the semantic segmentation performances.

In a popular work, Hoffman et al.\ \cite{hoffman2018} introduce the CyCADA framework, which relies on the CycleGAN \cite{cyclegan} image-to-image translator for pixel-level adaptation. First of all, a generative module, which is augmented with a semantic consistency constraint to avoid semantic alterations of input data, is trained to perform cross-domain image translation. Then, the generator responsible for the source-to-target mapping is applied on source images before they are provided to the predictor in a supervised training phase. Finally, a separate adversarial feature adaptation step is proposed. 
Unfortunately, the authors were not able to train the full framework with all the proposed modules,
due to hardware constraints hindering the insertion of a fully convolutional predictor (i.e., the segmentation network for semantic consistency) inside an already memory-demanding generative module comprising four neural networks.
Similarly to \cite{hoffman2018}, Chen et al.\ \cite{ChenL0H19} propose an extension of the CycleGAN framework by introducing a pair of feature domain discriminators and a couple of semantic segmentation networks.

\section{Architecture of the Proposed Approach}
\label{sec:method}

Our target is to train a semantic segmentation network  in a supervised way on synthetic data and then to adapt it in an unsupervised way to real world data. 
We assume to have access to synthetic (i.e., source) labeled images $(x_S,y_S)\in X_S \times Y_S$, as well as to real (i.e., target) unlabeled images $x_T \in X_T$. Due to dissimilar marginal and joint distributions over input and label spaces on both domains, deep models trained on source data struggle to generalize learned knowledge to the target space. 
\begin{figure*}
\centering
\includegraphics[width=0.87\textwidth]{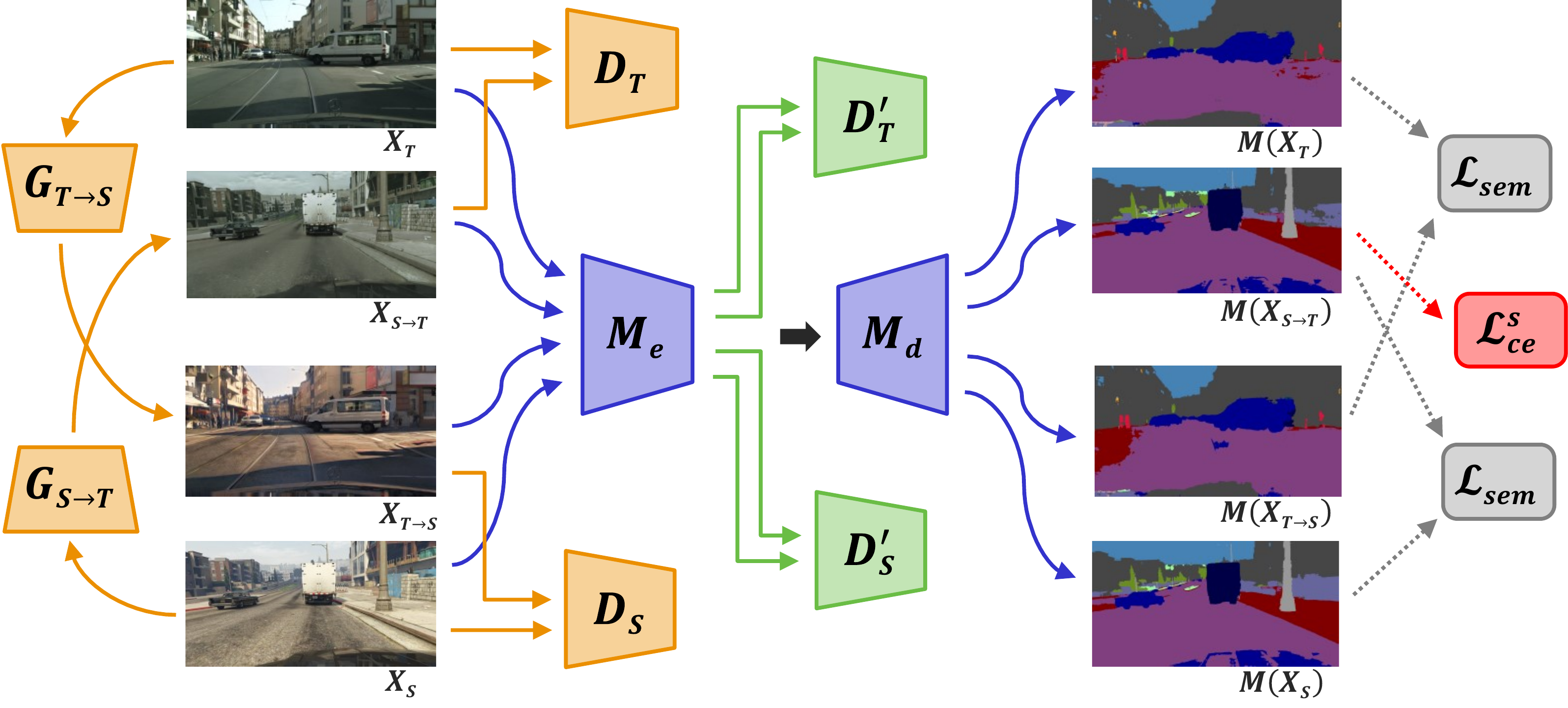}
\caption{Architecture of the proposed framework. 
Yellow blocks correspond to the CycleGAN framework for image-to-image translation. Original and translated scenes from both source and target sets are projected by the encoder to a latent space on which we apply an extra couple of domain discriminators (green blocks). Structural consistency on generated samples is enforced by the cycle-consistency constraint, whereas semantic uniformity throughout image mapping is promoted by the semantic loss. The segmentation network is reported in blue.
}
\label{fig:scheme}
\end{figure*}
To address the effect of domain discrepancy, we resort to a generative approach. We employ an adversarial framework to learn an image-level mapping between source and target spaces. The objective is to produce adapted source images that resemble target ones, while preserving the ground-truth information at our disposal. In this way, we can introduce a form of target supervision by exploiting target-like annotated source images to train the segmentation network.

Figure \ref{fig:scheme} shows the architecture of the proposed  framework. As initial step, we adopt a generative approach to learn an image-level mapping for cross-domain image projection. This is achieved using  an adversarial learning scheme exploiting a pair of generator-discriminator couples. 
Meanwhile, a semantic segmentation network is included to enforce semantic consistency to the generative process. The objective is to perform realistic sample translations, while preserving the semantic structure as identified by the semantic classifier. 
An additional feature-level adaptation is further included, in the form of a pair of feature discriminators. The way they operate is analogous to their image-level counterparts, as they enforce a statistical alignment of source and target data representations. The key difference lies in the operating space, since they act over an intermediate feature representation produced by the segmentation network, rather than directly within the original image domain. 

Our approach is independent of the segmentation architecture and in general any semantic segmentation network can be used, however in our experiments we used the MobileNet-v2 network \cite{mobilenet, mobilenetv2} embedded inside the DeepLab-v3+ framework \cite{encoder}. 
The primary component of this widely utilized model is the depthwise separable convolution, a lightweight reinterpretation of the standard convolutional layer responsible for both the efficiency and the reduced weight of the architecture. 
In addition, inverted residual blocks in place of the standard residual connections further enhance model compactness and size shrinkage.

\subsection{Cycle Consistent Domain Adaptation}

The generative module is based on the CycleGAN framework \cite{cyclegan}.  
The source to target (direct) mapping $G_{S\shortrightarrow T}\!\!: X_S\! \rightarrow \! X_T$ and the target to source one (inverse mapping) $G_{T\shortrightarrow S}\!\!: X_T \! \rightarrow \! X_S$  are discovered by means of an adversarial competition exploiting a couple of discriminators $D_{S}$ and $D_{T}$. 
The role of the domain discriminators, following the original concept of GANs \cite{gan}, is to discriminate between \textit{real} images in their original form and \textit{fake} images, i.e., synthetic data subjected to the domain translation. 
We resort to the standard adversarial objectives at the image level to train separately each generator-discriminator couple:
\begin{equation}
\centering
  \begin{split}
	\mathcal{L}_{i,T} & \left(  G_{S \shortrightarrow T}, D_{T}, X_S, X_T  \right) \\ 
	&= \mathbb{E}_{x_T\sim X_T}  \left[ \log \left( D_{T}(x_T) \right) \right] \\
	& + \mathbb{E}_{x_S\sim X_S} \left[ \log (1-D_T(G_{S \shortrightarrow T}(x_S))) \right] 
	\end{split}
	\label{eq:gan1}
\end{equation}
\begin{equation}
\centering
  \begin{split}
	\mathcal{L}_{i,S} & \left(  G_{T \shortrightarrow S}, D_S, X_T, X_S  \right) \\
	&= \mathbb{E}_{x_S\sim X_S}  \left[ \log \left( D_S(x_S) \right) \right] \\
	& + \mathbb{E}_{x_T\sim X_T} \left[ \log (1-D_S(G_{T \shortrightarrow S}(x_T))) \right] 
	\end{split}
	\label{eq:gan2}
\end{equation}

With the aforementioned generative process, we are able to learn a joint source-target distribution starting from the marginal ones, which means we can reproduce the same images in both source and target styles. Unfortunately, since there are infinite joint distributions that match the available marginal ones, we are not guaranteed that the mapping functions we discover are preserving content structure and semantics. In other words, without any additional constraint the $G_{S\shortrightarrow T}$ projection could completely disrupt input source images, still producing new samples with target properties, but far from their original versions.
For this reason, we employ an additional loss term enforcing cycle-consistency:
\begin{equation}
\centering
	\begin{split}
	\mathcal{L}_{cycle} & \left(  G_{S \shortrightarrow T}, G_{T \shortrightarrow S}, X_S, X_T  \right) \\
	&= \mathbb{E}_{x_S\sim X_S} \left[ \| G_{T \shortrightarrow S} \left( G_{S \shortrightarrow T} \left(x_S\right) \right) - x_S \| _1\right] \\
	& + \mathbb{E}_{x_T\sim X_T}  \left[ \| G_{S \shortrightarrow T} \left( G_{T \shortrightarrow S} \left(x_T\right) \right) - x_T \| _1\right]
	\end{split}
\end{equation}
The reconstruction requirement provided by the cycle-consistency loss $\mathcal{L}_{cycle}$ should encourage the preservation of structural properties throughout translations, resulting in a realistic image generation that does not affect the semantic content.

The adversarial strategy we adopt for conditional image generation has proven to be suitable for color and texture changes, but not for more radical geometrical transformations \cite{gan}. This is positive for our target, since we are looking for a cross-domain projection that allows us to safely transfer ground truth information from an original image to its translated version.

\subsection{Enforcing the Semantic Segmentation Consistency across Domains} 

Following the the idea introduced in \cite{hoffman2018}, we embed the generative module in a task-specific domain adaptation framework. The adversarial architecture is followed by a  network  performing semantic segmentation on data from both source and target domains.
In our work, we will assume the usage of a fully convolutional network, that in our implementation is the MobileNet-v2 network. We will denote it with $M=[M_{e},M_{d}]$, where $M_{e}$ is the encoder part of the network, while $M_{d}$ is the decoder. $M$ is pre-trained on the source domain, before its application in the proposed framework.

Due to the lack of labeling data on the target domain, we can not perform supervised training on this domain. However, we introduce a further loss component to enforce the semantic consistency on the generative action: the segmentation network $M$ is supplied with both original and adapted versions of the same image and we measure the semantic discrepancies (i.e., the differences in the segmentation network output) introduced by the projection between domains. The error information is then propagated back through the classifier up to the generator, which is optimized in order to minimize semantic alteration (among other objectives).
We impose this semantic uniformity by means of a semantic consistency loss:
\begin{equation}
  \centering
  \begin{split}
	\mathcal{L}_{sem} & \left( G_{S \shortrightarrow T}, G_{T \shortrightarrow S}, M, X_S, X_T \right) \\ 
	&= \mathcal{L}_{ce} \left( M, G_{S \shortrightarrow T}(X_S), \rho (M(X_S)) \right) \\ 
	&+ \mathcal{L}_{ce} \left( M, G_{T \shortrightarrow S}(X_T), \rho (M(X_T)) \right)
	\end{split}
\end{equation}
where $\rho(M(X))$ is the argmax of the output of the semantic segmentation network and:
\begin{equation} 
\mathcal{L}_{ce}  \left( M, X, Y \right) \! = \!
 - \mathbb{E}_{(x, y)\sim (X, Y)} \! \sum_{h,w,c}{y^{(h,w,c)} \cdot \log(M(x)^{(h,w,c)}) }
\end{equation}
Notice that the cross-entropy loss $\mathcal{L}_{ce}$ is computed over segmentation maps  obtained by applying the \textit{argmax} function $\rho$ on semantic predictions $M(X)$, rather than over ground-truth labels.
Therefore, its effect is not to promote correct semantic predictions, but to force the generator to yield transformed images that are semantically identical to the original ones when viewed under the scope of $M$.

This scheme allows us to perform a measure of semantic distance in both domains, without resorting to ground-truth information. On the other side, $M$ is not a perfect predictor (and on the target domain has typically lower performances), therefore an excessive emphasis on this loss may cause undesired artifacts in the generative process.

\subsection{Feature Level Domain Adaptation}

Aiming at further improving our domain adaptation framework, we introduce an additional adversarial module to perform feature-level domain adaptation. 
The core idea is to replicate the adversarial strategy adopted for the image-to-image translation task, where the goal was a pixel-level distribution alignment. 
The new  objective is instead the adaptation of intermediate feature representations, i.e., to ensure the proper adaptation at the level of the output of the encoder network $M_e$.
 The goal now is to make generated images from one domain appear statistically identical to original images from the other domain when looking at their projections in a latent space spanned by the segmentation network. More specifically, we add a couple of feature discriminators and feed them with activations from the output of the MobileNet's feature extractor $M_e$. The adversarial game, then, takes place between a couple of feature discriminators and the joint action of the two original generators together with the encoder of the segmentation network. The feature-level adversarial objectives are the following: 
\begin{equation}
\centering
  \begin{split}
	\mathcal{L}_{f,T} & (M_e \circ G_{S \shortrightarrow T}, D'_T  ) \\ 
	&= \mathbb{E}_{x_T\sim X_T}  [ \log ( D'_T(M_e(x_T)) ) ] \\
	& + \mathbb{E}_{x_S\sim X_S} [ \log (1-D'_T(M_e(G_{S \shortrightarrow T}(x_S)))) ] 
	\end{split}
	\label{eq:ganf1}
\end{equation}
\begin{equation}
\centering
  \begin{split}
	\mathcal{L}_{f,S} & ( M_e \circ G_{T \shortrightarrow S}, D'_S  ) \\ 
	&= \mathbb{E}_{x_S\sim X_S}  [ \log ( D'_S(M_e(x_S)) ) ] \\
	& + \mathbb{E}_{x_T\sim X_T} [ \log (1-D'_S(M_e(G_{T \shortrightarrow S}(x_T)))) ]  
	\end{split}
	\label{eq:ganf2}
\end{equation}
Where $D'_S$ and $D'_T$ are the feature discriminators working on data from the source and target domain respectively. 

Combining together all the different losses, the full objective becomes: 
\begin{equation}
\begin{split}
 \mathcal{L}_{tot} = ( \mathcal{L}_{i,S} + \mathcal{L}_{i,T}) + \lambda_{feat}\cdot(\mathcal{L}_{f,S}+ \mathcal{L}_{f,T}) \\
+ \lambda_{cycle}\cdot\mathcal{L}_{cycle}  + \lambda_{sem}\cdot\mathcal{L}_{sem} + \lambda_{ce}\cdot\mathcal{L}_{ce}^{s}
\end{split} 
\end{equation} 
We also denote $\mathcal{L}_{ce}^{s}=\mathcal{L}_{ce}(M, G_{S \shortrightarrow T}(X_S), Y_S)$ the standard cross-entropy loss used to train supervisedly $M$ on source adapted data. 
The framework optimization then can be expressed as a min-max problem:
\begin{equation}
 \min_{G_{S \shortrightarrow T},G_{T \shortrightarrow S},M} \max_{D_S,D_T,D'_S,D'_T} \mathcal{L}_{tot}. 
\end{equation}
As a result, we have access to a pair of image-to-image mappings capable of translating images across domains, while, at the same time, making generated samples statistically indistinguishable from true ones when projected into the feature space defined by the segmentation network. Additionally, the semantic segmentation network $M$ is adapted to work on the target data in an unsupervised way (without using target ground truth) thanks to the loss $\mathcal{L}_{ce}^{s}$. Since all the components are simultaneously trained, when improving the image-to-image mappings also the adaptation of $M$ improves.
For a more stable training and to avoid saturation effects \cite{cyclegan}, the logarithm within adversarial losses is replaced by a L2-norm operator and the objectives are split into separate terms for generators and discriminators individual optimization.

\section{Implementation and training of the proposed adversarial model}
\label{sec:training}

\noindent\textbf{Network implementation.}\hspace{0.8mm} 
The image-level generators and  discriminators (i.e., $G_{S \rightarrow T}$, $G_{T \rightarrow S}$, $D_S$ and $D_T$) are based on the network architectures introduced in \cite{cyclegan}.
 The generators are composed of stride-2 convolutions, residual blocks and fractionally strided convolutions to recover input dimensionality. Discriminators are fully convolutional networks as well, made by the cascade of 5 stride-2 convolutional layers. 
To avoid excessive size compression, we employ the same structure for feature-level discriminators ($D_S^f$ and $D_T^f$) as well, but we modify the stride value to 1 for all layers.
As concerns the segmentation network $M$, we employ the DeepLab-v3+ \cite{encoder} with the MobileNet-v2 \cite{mobilenetv2} as backbone. We select an output stride of 16 for the feature extractor, whereas for the ASPP block we choose atrous rates of 6, 12 and 18 as suggested by  \cite{encoder}.\\ 

\noindent\textbf{Training details.}\hspace{0.8mm} 
Differently from \cite{hoffman2018}, we train our framework in a single shot, so that all the networks are simultaneously optimized according to $\mathcal{L}_{tot}$. 
We initialize the weights of the semantic predictor from its pre-trained version on the Pascal VOC dataset \cite{Everingham2010, mobile_weights}. 
We then fine-tune it on the source domain for $90$K steps using the standard cross entropy loss before the actual optimization of the adaptation framework. 
All the other networks are instead trained from scratch. We use the Adam optimizer \cite{adam} to train all the components of the proposed approach.

After the initialization of the segmentation network, we train our model for a total of $80$K iterations with a single NVDIA GeForce GTX 1080 Ti. 
The segmentation network is kept fixed for the first $20$K steps, until the generators start performing acceptable translations, then we train all the various components together. 
The large amount of model parameters and the high resolution images from source and target sets prevents us from using full size data for training purposes. 
To overcome this issue, we extract random patches of $600\times600$ pixels from training samples, which were previously resized to have a 
predefined width 
and the original aspect ratio, and we use them as model inputs.

Concerning the parameters, we experimentally set the terms balancing the various components in $\mathcal{L}_{tot}$ to $\lambda_{cycle} = 20$, $\lambda_{sem} = 0.1$, $\lambda_{feat} = 10^{-4}$ and $\lambda_{ce} = 1$.
For the training of the segmentation network we set $\beta_1$ (the exponential decay rate for the first moment estimates) to $0.9$, and the weight decay to $4\times 10^{-5}$. The learning rate is subject to a polynomial decay of power $0.9$, and is decreased to $0$ from its initial value (respectively $1 \times 10^{-5}$ and $5 \times 10^{-6}$ for the adaptation from the GTA and SYNTHIA datasets). Moreover, we set a batch size of 5 when training the semantic predictor alone in the initial stage, while for full framework optimization, we reduce the batch size to 1 due to  memory constraints.
For the optimization of the image and feature adversarial models 
we use a small $\beta_1$ term of 0.5 as in \cite{cyclegan}. This makes the training process more unstable, but we notice no improvements by changing it. The learning rate for these modules is set to $2\times 10^{-4}$.

We developed our approach in TensorFlow and the code is publicly available at \url{https://lttm.dei.unipd.it/paper_data/UDAmobile/}.

\section{Experimental Results}
\label{sec:results}

In this section, we will briefly introduce the exploited datasets, then we will show some examples of the image translation module. Finally, we will show the results of the main task, i.e., semantic segmentation on real data, starting from two different synthetic datasets. In order to conclude, some ablation results are presented.

\subsection{Experimental Datasets}

In order to perform the experimental evaluation, we selected a source domain with easily accessible labeled samples that can be obtained in large quantities, i.e., synthetic imagery. Synthetic data not only comes with essentially free annotations and a much lower collection cost than real data thanks to automatic generation, but also in principle provides a total control over the virtual environment in terms of point of view, illumination, objects inside the scene, etc.  
On the opposite side, we used real images as the target domain. 
Before going into details, notice that even if the experimental evaluation is performed on the synthetic to real adaptation, the proposed approach can be exploited in any domain adaptation task, not only in this specific setting.

We assume no ground-truth information is accessible in any form for real data (we used real labels only for evaluation of the results), as we are in a fully unsupervised scenario in this domain.
As concerns the specific datasets used to train and validate our framework, we choose the publicly accessible GTA5 \cite{Richter2016}, SYNTHIA \cite{ros2016} and Cityscapes \cite{Cordts2016} datasets.
They are built specifically to address semantic segmentation of urban scenes, which is of strategic importance in autonomous driving. 
The task is quite challenging, since multiple objects of different sizes with very different occurrence frequencies and various semantic categories have to be recognized with high confidence.

The \textbf{GTA5} dataset \cite{Richter2016} contains 24966 synthetic images rendered using the graphics engine of the popular video game \textit{Grand Theft Auto 5}. Its images are taken from the car perspective in the streets of US-like virtual cities and shows remarkable quality and very realistic scenes. We used $23966$ images for the supervised training and $1000$ images have been held-out for validation purposes. It is labeled using $19$ semantic classes which are directly comparable with the ones from the Cityscapes dataset.

The images in the \textbf{SYNTHIA} dataset \cite{ros2016} have been generated through an ad-hoc engine and represent many types of street scenes, acquired from different angles (not only from car drivers viewpoint but also from video-surveillance cameras, from pedestrians, etc...) in various illumination and weather conditions in European-style cities. The visual quality is lower than the one of the GTA5 dataset. In particular, we employed the \textit{SYNTHIA-RAND-CITYSCAPES} subset of the SYNTHIA dataset, which is comprised of $9400$ synthetic images of $1280\times760\mathrm{px}$ and has $16$ out of $19$ semantic classes comparable with the ones of the Cityscapes dataset. We used $9300$ images for the supervised training while $100$ images have been held-out for validation purposes.

The \textbf{Cityscapes} dataset \cite{Cordts2016} contains 2975 images of $2048\times1024\mathrm{px}$ acquired on the streets of different German cities. Only the semantic classes in common with the respective synthetic dataset used for the first supervised stage have been considered. No labels from this dataset have been used during the training procedure and the final results are reported on the original validation set of 500 images as done by all competing approaches.

\subsection{Image Domain Translation}
\label{sec:img_translation}

The first module in our framework is  the image-to-image adaptation in the image space between the synthetic datasets and the real one (Cityscapes) and viceversa. We visualize such cyclic translation in Figure \ref{fig:domain_transfer}, where we report the original, adapted and reconstructed images in output from our model for each of the four different considered scenarios. We show both the results when starting from synthetic data and when starting from real scenes. 

The most obvious and noticeable difference lies in the shift of colors between real and synthetic images. Synthetic imagery, indeed, are characterized by more vivid colors than the real counterparts hence the adaptation framework needs to compensate for this issue as can be verified in all the proposed qualitative results.
Additionally, the textures of some regions (such as the road or the sky) of the images are completely different between the considered domains. In GTA5 and especially in SYNTHIA (where the road texture is not realistic at all) the road is less uniform than the one present in the Cityscapes dataset, so we could observe that our adaptation framework compensate this aspect. In particular, the model adds some textures on the road when converting an image to the synthetic dataset and it removes such textures when dealing with the opposite task. The sky, instead, tends to appear more blue in the synthetic imagery rather than in the real ones, where it appears more grayish.

Beside those changes in the appearance of the images, we could notice that in general the semantic content and the geometrical structure is preserved unaltered. The objects do not disappear and they do not change position, as we expect. This is ensured by a combination of factors such as the semantic loss, the cycle-consistency loss and the adaptation loss at the feature level which aims at preserving the extraction of similar features from the same objects even if they belong to two different image-spaces.

Furthermore, when moving from synthetic to real domain, in certain pictures we can observe that our model tends to add an ornament to the bottom of the images (e.g., in the second row from GTA5 to Cityscapes). Although this is irrelevant for the semantic segmentation task, since we exclude the car on which the camera is mounted, we argue that the image-space adaptation is leading to reasonable outcomes because it tries to replicate the trademark of the car used for the Cityscapes data acquisition.

A few artifacts are present especially in the sky region where the model tends to add some shadows coming from clouds or buildings present in other images: this can be noticed in the first and third adapted images from the GTA5 dataset. However, it is noticeable that the cycle-consistency helps the model to recover the exact appearance of the sky of such images.

\newcommand{\tablecenter}{\raisebox{-0.12cm}}

\newcommand{\imgsize}{28.55mm}
\begin{figure*}[htbp]
\setlength{\tabcolsep}{1pt} 
\renewcommand{\arraystretch}{1.2}
\
\centering
\begin{subfigure}[htbp]{2\textwidth}
\begin{tabular}{c|>{\hspace{0.2pc}}ccc<{\hspace{0.2pc}}|>{\hspace{0.2pc}}ccc<{\hspace{0.2pc}}|}
  
   \multicolumn{1}{c}{}&\multicolumn{3}{c}{Synthetic $\rightarrow$ Real$\rightarrow$ Synthetic} & \multicolumn{3}{c}{ Real $\rightarrow$ Synthetic $\rightarrow$  Real} \\
   \cmidrule{2-7}
   \multirow{2}{*}{\rotatebox{90}{ GTA5 $\leftrightarrow$ Cityscapes \hspace{-1ex}}} & 
   \tablecenter{\includegraphics[width=\imgsize]{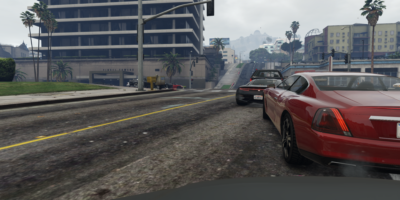}} &
   \tablecenter{\includegraphics[width=\imgsize]{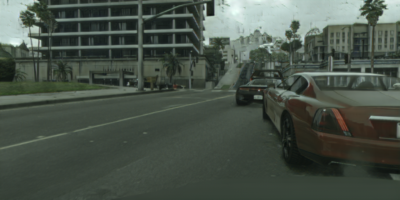}} & 
   \tablecenter{\includegraphics[width=\imgsize]{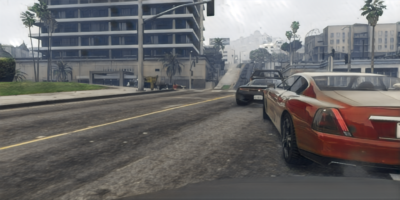}} & 
   \tablecenter{\includegraphics[width=\imgsize]{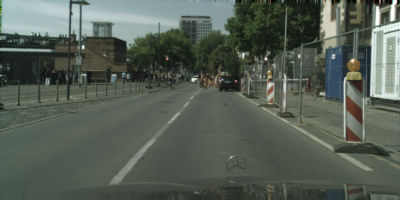}} & 
   \tablecenter{\includegraphics[width=\imgsize]{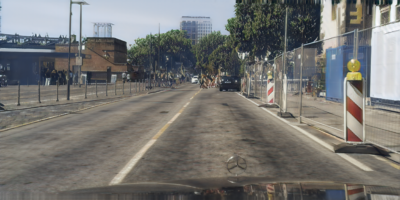}} & 
   \tablecenter{\includegraphics[width=\imgsize]{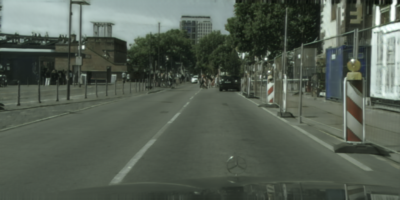}} \\
  
   &
   \tablecenter{\includegraphics[width=\imgsize]{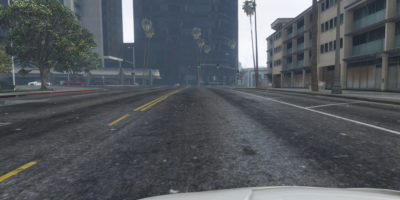}} &
   \tablecenter{\includegraphics[width=\imgsize]{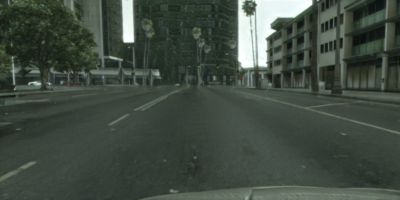}} & 
   \tablecenter{\includegraphics[width=\imgsize]{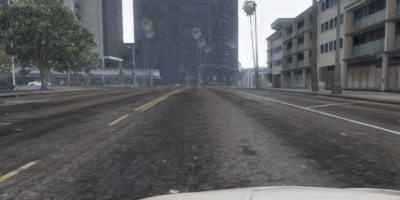}} & 
   \tablecenter{\includegraphics[width=\imgsize]{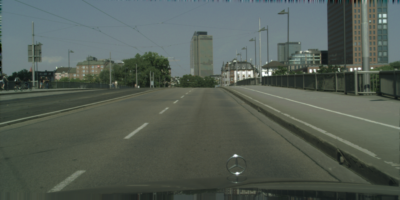}} & 
   \tablecenter{\includegraphics[width=\imgsize]{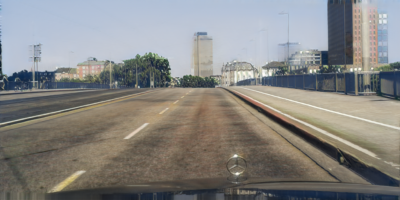}} & 
   \tablecenter{\includegraphics[width=\imgsize]{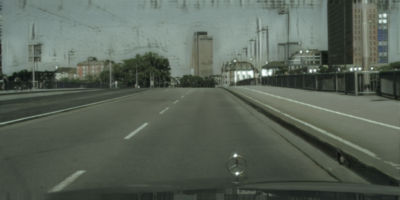}} \\
  
   &
   \tablecenter{\includegraphics[width=\imgsize]{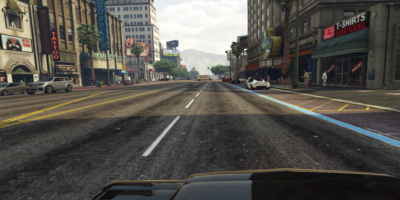}} &
   \tablecenter{\includegraphics[width=\imgsize]{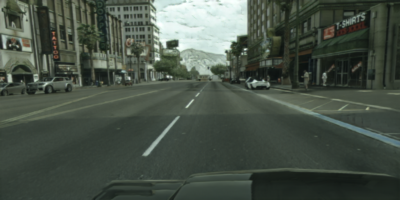}} & 
   \tablecenter{\includegraphics[width=\imgsize]{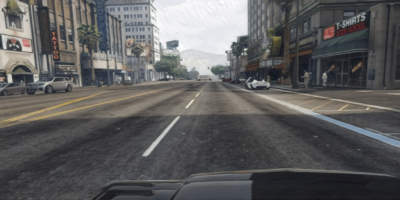}} & 
   \tablecenter{\includegraphics[width=\imgsize]{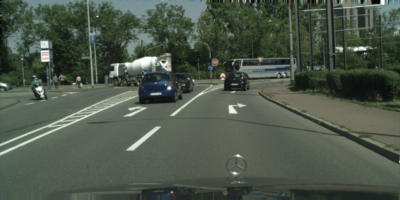}} & 
   \tablecenter{\includegraphics[width=\imgsize]{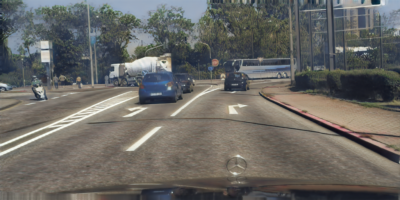}} & 
   \tablecenter{\includegraphics[width=\imgsize]{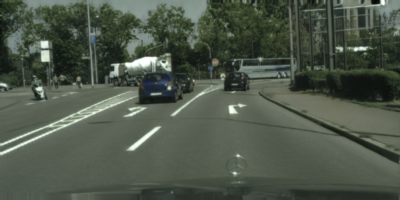}} \\
  \cmidrule{1-7} 
  
   \multirow{3}{*}{\rotatebox{90}{ SYNTHIA $\leftrightarrow$ Cityscapes \hspace{-3.5ex}}} &
   \tablecenter{\includegraphics[width=\imgsize]{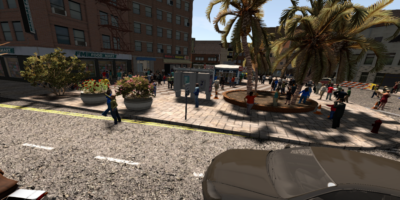}} &
   \tablecenter{\includegraphics[width=\imgsize]{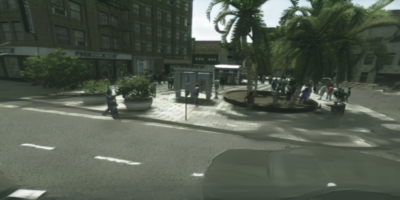}} & 
   \tablecenter{\includegraphics[width=\imgsize]{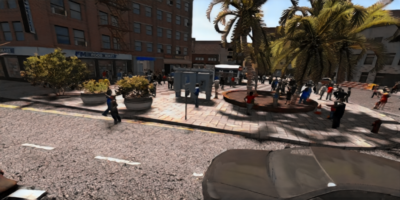}} & 
   \tablecenter{\includegraphics[width=\imgsize]{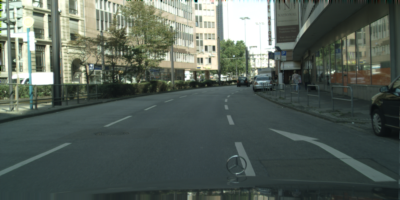}} & 
   \tablecenter{\includegraphics[width=\imgsize]{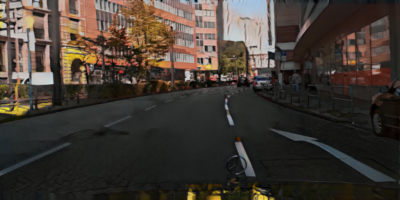}}& 
   \tablecenter{\includegraphics[width=\imgsize]{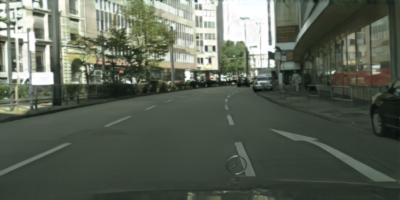}} \\

   &
  \tablecenter{\includegraphics[width=\imgsize]{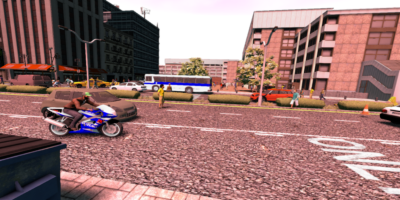}} &
   \tablecenter{\includegraphics[width=\imgsize]{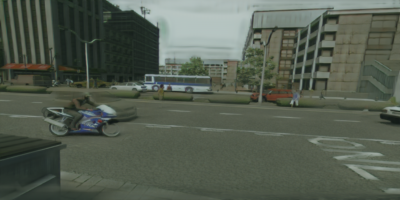}} & 
   \tablecenter{\includegraphics[width=\imgsize]{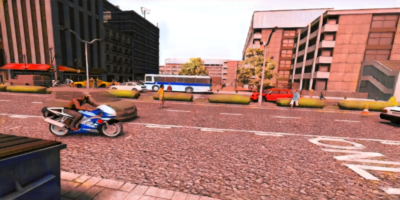}} & 
   \tablecenter{\includegraphics[width=\imgsize]{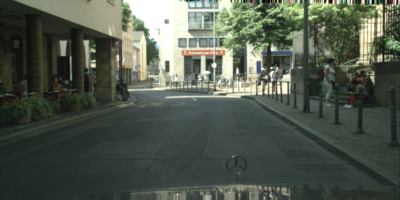}} & 
   \tablecenter{\includegraphics[width=\imgsize]{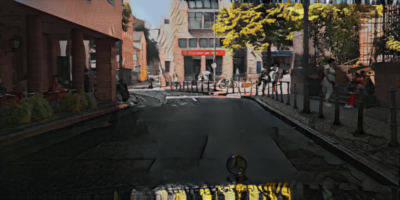}}& 
   \tablecenter{\includegraphics[width=\imgsize]{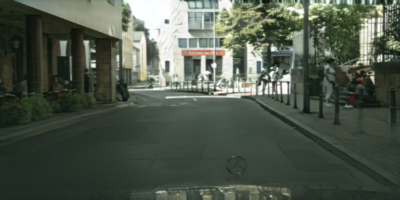}} \\
  
 &
  \tablecenter{\includegraphics[width=\imgsize]{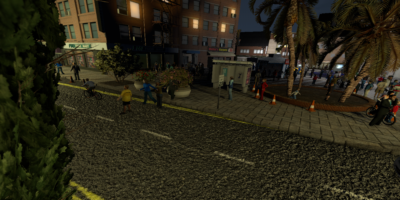}} &
   \tablecenter{\includegraphics[width=\imgsize]{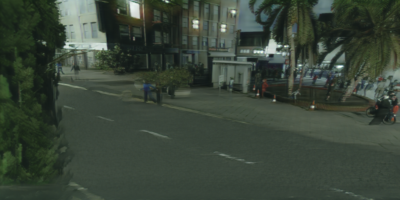}} & 
   \tablecenter{\includegraphics[width=\imgsize]{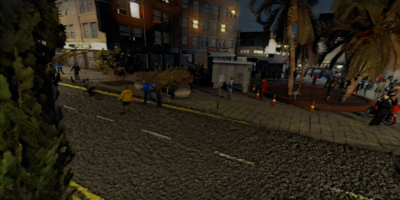}} & 
   \tablecenter{\includegraphics[width=\imgsize]{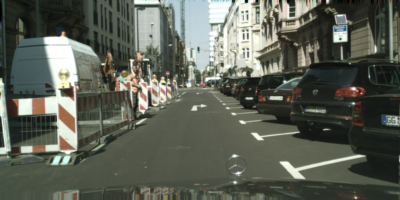}} & 
   \tablecenter{\includegraphics[width=\imgsize]{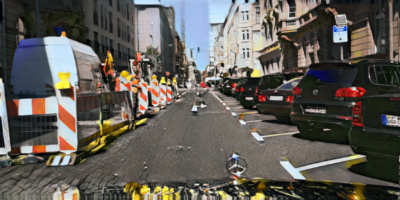}}& 
   \tablecenter{\includegraphics[width=\imgsize]{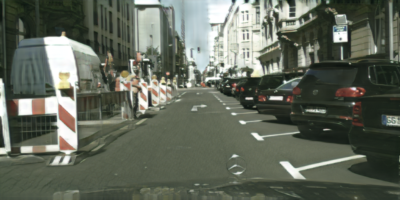}} \\
 \cmidrule{2-7}
 \multicolumn{1}{c}{} & original & adapted & \multicolumn{1}{c}{reconstructed} & original & adapted & \multicolumn{1}{c}{reconstructed} \\
  
\end{tabular}
\end{subfigure}
\centering
\caption{Examples of image translations (adaptation and reconstruction of the original image) in the four different cases considered. In the first quadrant (up-left of $3\times 3$ images) we move from the GTA5 dataset to the adapted images in the domain of the Cityscapes dataset to the reconstructed images in the GTA5 domain. The second quadrant (up-right) shows respectively: the starting Cityscapes images, their translation to the GTA dataset and the reconstructed images in the domain of the Cityscapes images. The third (down-left) and the fourth (down-right) quadrant are analogous to the first and the second ones using the SYNTHIA dataset in place of GTA5.}
\label{fig:domain_transfer}
\end{figure*}

\begin{table*}[htbp]
\setlength{\tabcolsep}{1.6pt}
\centering
\begin{tabular}{|c|ccccccccccccccccccc|c|}
\hline
 & \rotatebox{90}{road} &  \rotatebox{90}{sidewalk} &  \rotatebox{90}{building} &  \rotatebox{90}{wall} &\rotatebox{90}{fence} & \rotatebox{90}{pole} & \rotatebox{90}{t light} 
  &\rotatebox{90}{t sign} & \rotatebox{90}{veg} & \rotatebox{90}{terrain} & \rotatebox{90}{sky} & \rotatebox{90}{person}& \rotatebox{90}{rider} & \rotatebox{90}{car} 
  & \rotatebox{90}{truck} & \rotatebox{90}{bus} & \rotatebox{90}{train} &  \rotatebox{90}{mbike} & \rotatebox{90}{bike} & \rotatebox{90}{mean} \\
 \hline
Source only & 23.1 & 13.1 & 42.6 &  2.3 & 13.9 &  5.0 & 10.3 &  8.0 & 68.6 &  6.7 & 24.5 & 40.8 &  0.3 & 48.1 &  9.4 & 16.3 &  0.0 &  0.0 &  0.0 & 17.5 \\
Ours (full) & \textbf{87.6} & \textbf{36.7} & \textbf{83.5} & \textbf{29.1} & \textbf{17.8} & \textbf{33.6} & 24.3 & \textbf{35.2} & \textbf{83.1} & \textbf{28.9} & \textbf{76.3} & \textbf{59.1} & \textbf{14.0} & \textbf{85.9} & \textbf{25.4} & \textbf{29.4} &  \textbf{2.6} & \textbf{19.5} &  \textbf{9.3} & \textbf{41.1} \\
\hline
CycleGAN \cite{cyclegan} & 84.9 & 36.4 & 74.3 & 12.9 &  7.1 & 23.6 &  7.9 & 19.9 & 60.2 & 13.4 & 45.8 & 46.8 &  5.4 & 72.4 & 18.0 & 22.3 &  0.8 &  3.2 &  0.5 & 29.3 \\
CyCADA \cite{hoffman2018} & 83.8 & 35.3 & 80.4 & 20.7 & 15.7 & 28.4 & \textbf{27.0} & 24.8 & 80.2 & 23.1 & 69.0 & 56.6 & 11.5 & 80.8 & 23.4 & 27.0 &  2.4 & 12.4 &  5.2 & 37.3 \\ 
\hline
Oracle      & 97.7 & 81.9 & 91.0 & 47.6 & 50.1 & 58.4 & 62.3 & 73.4 & 91.4 & 59.8 & 94.3 & 77.2 & 50.5 & 93.2 & 59.2 & 74.8 & 55.8 & 49.5 & 73.0 & 70.6\\

\hline
\end{tabular}
\caption{mIoU on the different classes of the Cityscapes validation set. The approaches have been trained in a supervised way on the GTA5 dataset and the unsupervised domain adaptation has been performed using the Cityscapes training set. The mean and per class highest results have been highlighted in bold.}
\label{tab:GTA}

\vspace{0.5cm}

\setlength{\tabcolsep}{3.6pt}
\centering
\begin{tabular}{|c|cccccccccccccccc|c|}
\hline
 & \rotatebox{90}{road} &  \rotatebox{90}{sidewalk} &  \rotatebox{90}{building} &  \rotatebox{90}{wall} &\rotatebox{90}{fence} & \rotatebox{90}{pole} & \rotatebox{90}{t light} 
  &\rotatebox{90}{t sign} & \rotatebox{90}{veg} & \rotatebox{90}{sky} & \rotatebox{90}{person}& \rotatebox{90}{rider} & \rotatebox{90}{car} 
 & \rotatebox{90}{bus} &  \rotatebox{90}{mbike} & \rotatebox{90}{bike} & \rotatebox{90}{mean} \\
 \hline
Source only &  1.4 & 10.6 & 29.1 &  1.0 &  0.0 & 17.2 &  2.0 &  3.6 & 68.5 & 65.0 & 42.3 &  0.1 & 41.7 &  8.2 &  0.0 &  0.5 & 18.2 \\
Ours (full) & \textbf{53.8} & 21.3 & \textbf{69.4} &  3.7 &  \textbf{0.1} & \textbf{31.6} &  3.5 & \textbf{12.6} & \textbf{77.5} & \textbf{75.2} & \textbf{51.9} & \textbf{13.2} & \textbf{64.1} & \textbf{15.9} & \textbf{10.8} & \textbf{16.7} & \textbf{32.6} \\
\hline
CycleGAN \cite{cyclegan} & 42.9 & \textbf{26.7} & 44.6 &  0.5 &  \textbf{0.1} & 29.2 &  3.5 &  5.8 & 67.1 & 70.8 & 46.0 &  3.4 & 32.8 &  7.0 &  2.4 &  3.7 & 24.2 \\
CyCADA \cite{hoffman2018} & 30.3 & 15.8 & 64.0 &  \textbf{5.9} &  0.0 & 30.6 &  \textbf{3.8} & 10.4 & 76.4 & 73.0 & 42.9 &  4.9 & 54.3 & 15.0 &  3.0 &  8.9 & 27.5 \\
\hline
Oracle & 97.8 & 83.7 & 91.2 & 47.7 & 49.7 & 58.8 & 63.0 & 73.9 & 92.4 & 94.4 & 77.9 & 53.7 & 94.1 & 80.5 & 44.7 & 73.5 & 73.6\\
\hline
\end{tabular}
\caption{mIoU on the different classes of the Cityscapes validation set. The approaches have been trained in a supervised way on the SYNTHIA dataset and the unsupervised domain adaptation has been performed using the Cityscapes training set. The mean and per class highest results have been highlighted in bold.}
\label{tab:SYNTHIA}
\end{table*}

\subsection{Domain adaptation from the GTA5 dataset}

The first set of experiments regards the unsupervised adaptation of the semantic segmentation network $M$ to the Cityscapes dataset after an initial stage of supervised training on the GTA5 dataset. 
To evaluate the adaptation performance of our framework we computed the mean Intersection over Union (mIoU) between model predictions and the relative ground truth label maps for the scenes in the Cityscapes validation set.

The results of these synthetic to real adaptation experiments are summarized in Table \ref{tab:GTA}.  
The baseline approach, i.e., the supervised training of the semantic segmentation on the GTA5 (source) dataset followed by testing on Cityscapes without any adaptation, leads to a very low mIoU of $17.5 \%$ (first row). 
When compared to the training of the same network on the target (Cityscapes) dataset 
(last row, denoted as \textit{Oracle}), the huge difference reveals the struggle of the predictor to overcome the statistical discrepancy between the source and target domains. 

Our unsupervised domain adaptation method brings a huge improvement over the na\"ive source only approach, reaching a mIoU of $41.1 \%$ when employed in its full extent (5th row), with a performance boost of $23.6 \%$ over the baseline.
Moreover, the accuracy enhancement is well distributed over all the semantic classes, from the most common ones (e.g.\ \textit{road}, \textit{building}, \textit{sky}), to the less frequent categories (e.g.\ \textit{train}, \textit{motobike}, \textit{bike}).
In particular notice that some of the less frequent are never recognized when no adaptation is performed while the proposed approach allows to detect them even if they remain very challenging. 
This proves the effectiveness of our model in mitigating the statistical discrepancy through a combined feature and pixel level alignment. 

We compared the results we obtained on the Cityscapes validation set with two approaches of the same family: namely, CycleGAN \cite{cyclegan} and CyCADA \cite{hoffman2018}. Those are very well known approaches and also represent the starting point for some architectural design choices we made. 
For comparison purposes, we used the CycleGAN implementation from \cite{cyclegan_implementation} and we implemented the framework of \cite{hoffman2018} from scratch and inserted in it the same segmentation network we used in our tests (i.e.\ DeepLabV3+ \cite{encoder} with MobileNet-v2 \cite{mobilenetv2} as backbone). As regards the training details of the compared methodologies, we employed the standard training procedures proposed in the respective papers: for CycleGAN \cite{cyclegan} we train the model with rescaled images, while for CyCADA \cite{hoffman2018} we train the model extracting random patches after a rescaling operation.

From Tables \ref{tab:GTA} and \ref{tab:SYNTHIA} we can appreciate that our method is able to outperform CyCADA by about $4 \%$ and $5\%$ respectively, which is uniformly distributed among classes: notice how our approach is the best on 18 out of 19 classes (it is outperformed only by \cite{hoffman2018} on the traffic light class). This has to be mostly attributed to the feature alignment process which directly influences the generative action and also to the simultaneous optimization of all framework components. Additionally, we could observe that CycleGAN has a much lower accuracy and lies in between the training made only on source data and our full method. Indeed, it consists of a simpler framework where only the image translation based on the cycle-consistency constraint is present.

Figure \ref{fig:qual_res} displays some qualitative results in terms of semantic prediction maps for different adaptation strategies. The first and second columns include the original Cityscapes RGB images and their corresponding label maps, while the last 3 columns show the segmentation outputs with no adaptation (i.e.\ \textit{source only}), using the approach of \cite{hoffman2018} and with the proposed unsupervised domain adaptation strategy.

The accuracy enhancement introduced by our adaptation strategy can be appreciated from the visual results in Figure \ref{fig:qual_res}. 
The first 4 rows   show the remarkable improvement achieved over the \textit{source only} baseline when resorting to the GTA5 dataset as source domain. The effect of the adaptation is to boost the detection capability of the predictor $M$, as it manages to obtain a more accurate understanding of the input target scene, both in terms of correct categorization and precise spatial identification of the different semantic entities. All semantic classes happen to highly benefit from the adaptation, leading to generally better semantic predictions. Anyway some categories exhibit more noticeable improvements, such as the \textit{road} and \textit{sky} classes as it is possible to see on the predicted maps in rows 1, 2 and 4.
\\
Column 4 of Figure \ref{fig:qual_res} shows the output of \cite{hoffman2018}: 
our method outperforms \cite{hoffman2018} on the majority of the semantic classes, with a particularly noticeable improvement on some of them. 
For instance, our approach in general leads to a more precise segmentation of the upper portion of target scenes, mainly involving the \textit{sky} and \textit{building} classes. It is possible to notice how the approach of \cite{hoffman2018} sometimes confuses part of \textit{sky} as \textit{terrain} or \textit{building} (rows 1, 2 and 4).
As previously stated, this is due to the dataset bias occurring between source and target domains in terms of color shift and change of texture, making the detection of semantic components quite hard to achieve, 
especially when the classification involves different categories (such as \textit{sky} and \textit{building}) sharing a common appearance on the source and target domains. 
Furthermore, we observe that less frequent classes corresponding to small image details (e.g.,\ \textit{traffic} \textit{sign}) are more precisely localized, as well as more common categories lacking an always definite semantic categorization  
(e.g.\ \textit{wall} and \textit{sidewalk}). Once again, this remarks the capability of our adaptation framework to address domain discrepancy with the combined pixel and feature level alignment.

\subsection{Domain adaptation from the  SYNTHIA dataset}

In the second set of  experiments we changed the source domain, replacing the GTA5 dataset with the SYNTHIA one. 
Table \ref{tab:SYNTHIA} shows the numerical results we got from the evaluation.
As before, we started by training the semantic segmentation network on synthetic data and measuring its performance on the Cityscapes validation set, which we employ as a baseline (first row). 
The accuracy ($18.2 \%$) happens to be slightly higher than in the first scenario, even if the adaptation task is more challenging due to a wider dataset bias. 
Anyway, the huge performance gap w.r.t. the target-based supervised optimization still persists, leaving large room for improvement. 

Our adaptation framework managed to reduce the domain discrepancy quite successfully boosting the mIoU up to $32.6 \%$ from the original $18.2 \%$ of the baseline, with an improvement of almost $15 \%$.
As for the  GTA5 to Cityscapes adaptation, the accuracies for all semantic categories benefit from the unsupervised adaptation. 
For example, road segmentation is strongly improved, starting from a mIoU for the road class of $1.4 \%$ (denoting basically no detection capability) and reaching a final accuracy of $53.8 \%$. 
This class is particularly interesting since the reason for the low accuracy lies in the rather unrealistic synthetic road texture of SYNTHIA images, whose semantic attributes are not easily generalizable to the Cityscapes dataset.
To that end, our framework introduces a substantial distribution alignment both at pixel level (as discussed in Section \ref{sec:img_translation}) and inside the intermediate feature space.
We once again compared the adaptation performance of our model with the one of CyCADA \cite{hoffman2018} in the third row. Our framework is more  effective than the competitor, with an average mIoU increase of $5 \%$, shared by the majority of the classes. For instance, the road accuracy is significantly  enhanced by our model, proving the better capability of reducing domain discrepancy. Again, the CycleGAN \cite{cyclegan} approach has intermediate results between the source only approach and CyCADA \cite{hoffman2018}.
\\

Some qualitative results for the adaptation from the SYNTHIA dataset are shown last 4 rows of Figure \ref{fig:qual_res}. 
They confirm the numerical evaluation: the strong diversity between the Cityscapes and SYNTHIA datasets negatively affects the semantic understanding of the predictor, as clearly visible in the third column showing the outputs of the baseline approach with no adaptation. Even common classes such as \textit{road} and \textit{building} suffer from quite flawed semantic detection leading to almost no understanding of target scene structure, with a worse performance when compared to the GTA5 scenario. This has to be ascribed to the poor realism of synthetic images and to the variable view point of SYNTHIA scenes, which are captured from multiple camera angles and not exclusively from a car perspective as for the Cityscapes images. 
The adaptation successfully mitigates the domain discrepancy providing the predictor with an enhanced perception of the semantic morphology of target inputs. For example, the \textit{road}, which without adaption is incorrectly classified as \textit{building}, probably due to its quite unrealistic texture on source images, after the adaptation is detected with a much greater accuracy.

Furthermore, our approach shows some improvement also w.r.t. the method proposed in \cite{hoffman2018}. For example, the adaption following \cite{hoffman2018} struggles in the correct segmentation of \textit{road} and \textit{sidewalk} classes, which are easily mistaken one for the other, an issue greatly reduced in the maps of our approach in the last column. 
At the same time, semantic predictions exhibit a better detection accuracy on objects belonging to low frequency classes, such as \textit{motorbike} and \textit{bike}, highlighting an increased robustness of our method due to the combined pixel and feature level alignment and to the simultaneous optimization of all the network components.

\renewcommand{\imgsize}{34.1mm}
\renewcommand{\tablecenter}{\raisebox{-0.09cm}}
\begin{figure*}[htbp]
\centering
\begin{subfigure}[htbp]{\textwidth}
\hspace{-0.08cm}
\resizebox{1\textwidth}{!}{%
\begin{tabular}{cccccccccc}
\cellcolor[HTML]{804080}{\color[HTML]{FFFFFF} \textbf{road}} & \cellcolor[HTML]{F423E8}\textbf{sidewalk} & \cellcolor[HTML]{464646}{\color[HTML]{FFFFFF} \textbf{building}} & \cellcolor[HTML]{66669C}{\color[HTML]{FFFFFF} \textbf{wall}} & \cellcolor[HTML]{BE9999}\textbf{fence} & \cellcolor[HTML]{999999}\textbf{pole} & \cellcolor[HTML]{FAAA1E}\textbf{traffic light} & \cellcolor[HTML]{DCDC00}\textbf{traffic sign} &\cellcolor[HTML]{6B8E23} \textbf{vegetation} & \cellcolor[HTML]{98FB98}\textbf{terrain} \\ \cline{10-10} 
\cellcolor[HTML]{4682B4}\textbf{sky} & \cellcolor[HTML]{DC143C}{\color[HTML]{FFFFFF} \textbf{person}} & \cellcolor[HTML]{FF0000}{\color[HTML]{FFFFFF} \textbf{rider}} & \cellcolor[HTML]{00008E}{\color[HTML]{FFFFFF} \textbf{car}} & \cellcolor[HTML]{000046}{\color[HTML]{FFFFFF} \textbf{truck}} & \cellcolor[HTML]{003C64}{\color[HTML]{FFFFFF} \textbf{bus}} & \cellcolor[HTML]{005064}{\color[HTML]{FFFFFF} \textbf{train}} & \cellcolor[HTML]{0000E6}{\color[HTML]{FFFFFF} \textbf{motorcycle}} & \multicolumn{1}{c|}{\cellcolor[HTML]{770B20}{\color[HTML]{FFFFFF} \textbf{bicycle}}} & \multicolumn{1}{c|}{\textbf{unlabeled}} \\ \cline{10-10} 
\end{tabular}%
}
\vspace{0.1cm}
\end{subfigure}
\setlength{\tabcolsep}{1.5pt} 
\centering
\begin{subfigure}[htbp]{\textwidth}
\begin{tabular}{|c|c|ccccc}
\cline{1-2}
  
  \multirow{8}{*}{\rotatebox{90}{\hspace{-30ex}To Cityscapes}} & 
  \multirow{4}{*}{\rotatebox{90}{\hspace{-13ex}From GTA5}} &

   \tablecenter{\includegraphics[width=\imgsize]{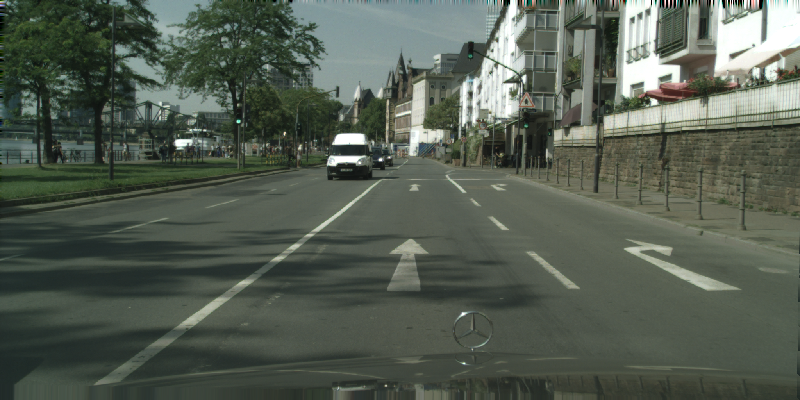}} &
   \tablecenter{\includegraphics[width=\imgsize]{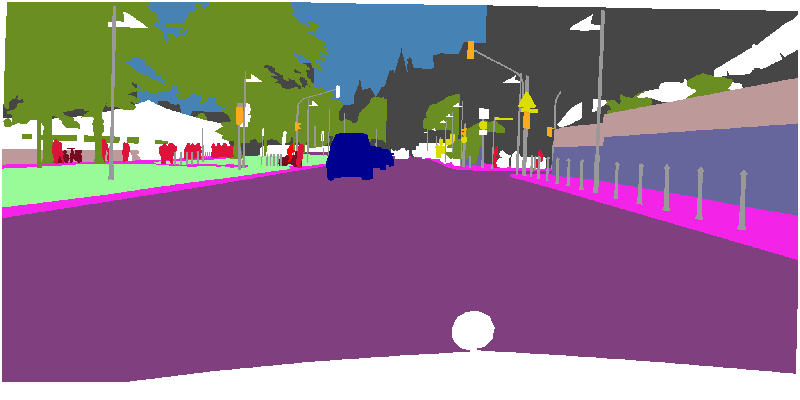}} & 
   \tablecenter{\includegraphics[width=\imgsize]{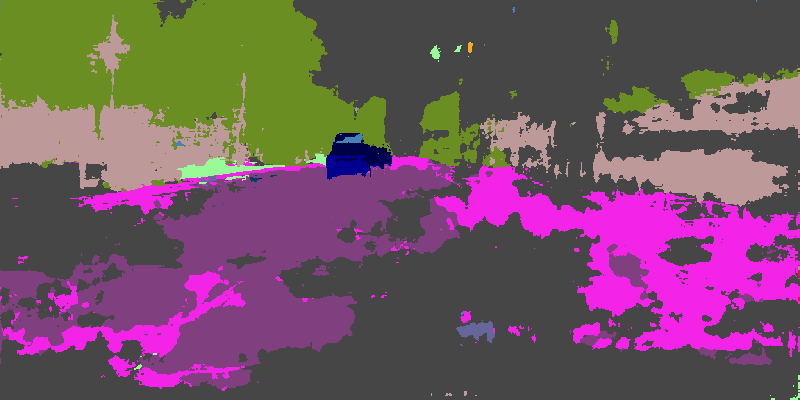}} & 
   \tablecenter{\includegraphics[width=\imgsize]{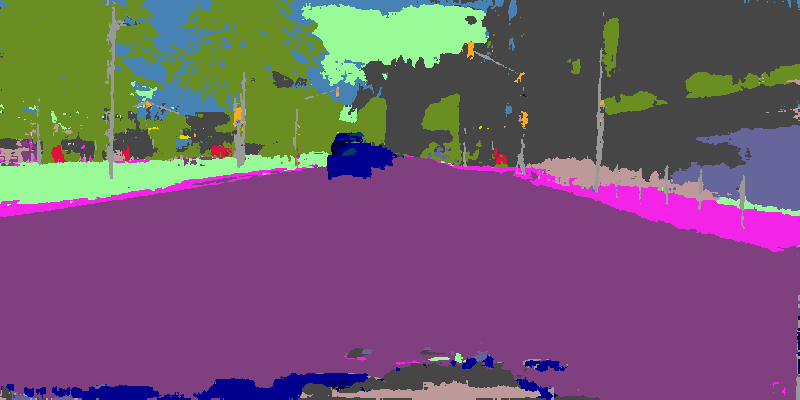}} & 
   \tablecenter{\includegraphics[width=\imgsize]{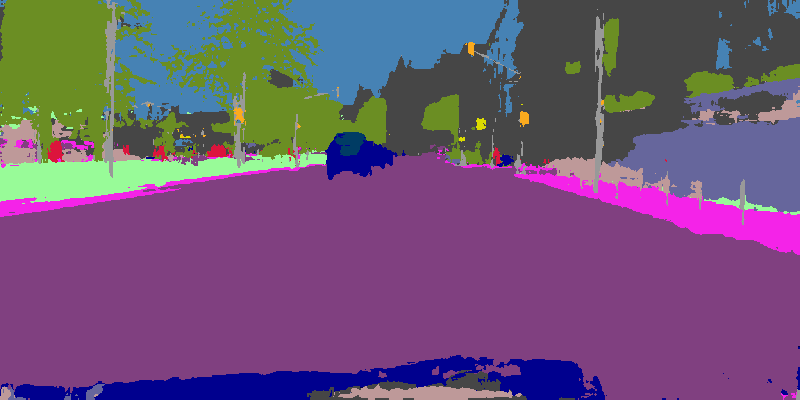}} \\
   & &  
   \tablecenter{\includegraphics[width=\imgsize]{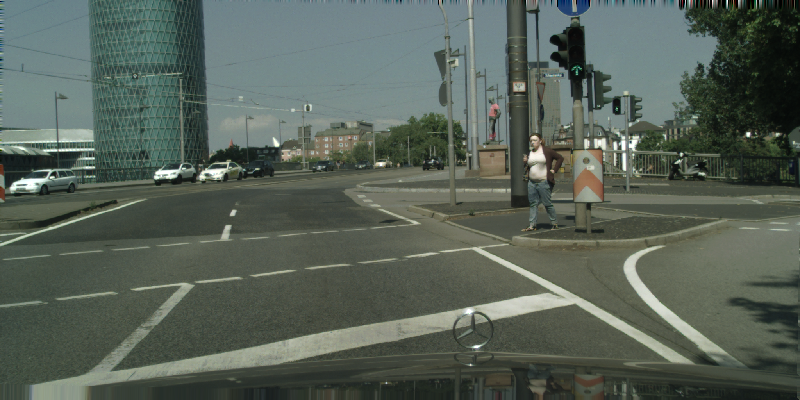}} &
   \tablecenter{\includegraphics[width=\imgsize]{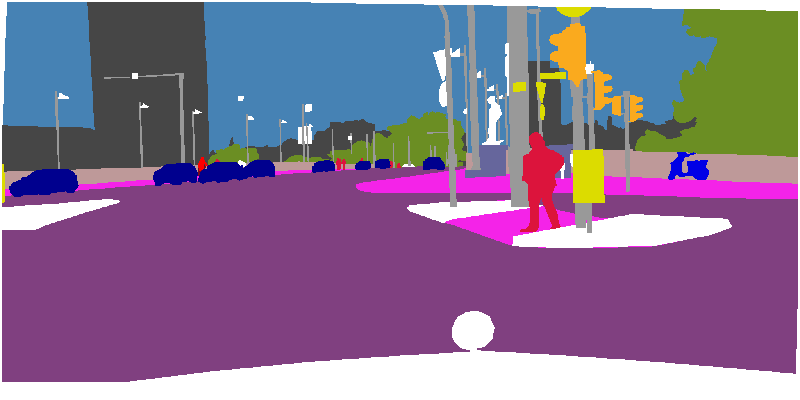}} & 
   \tablecenter{\includegraphics[width=\imgsize]{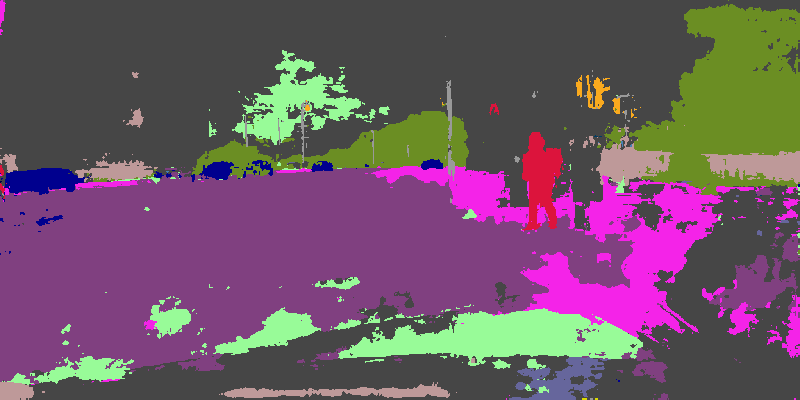}} & 
   \tablecenter{\includegraphics[width=\imgsize]{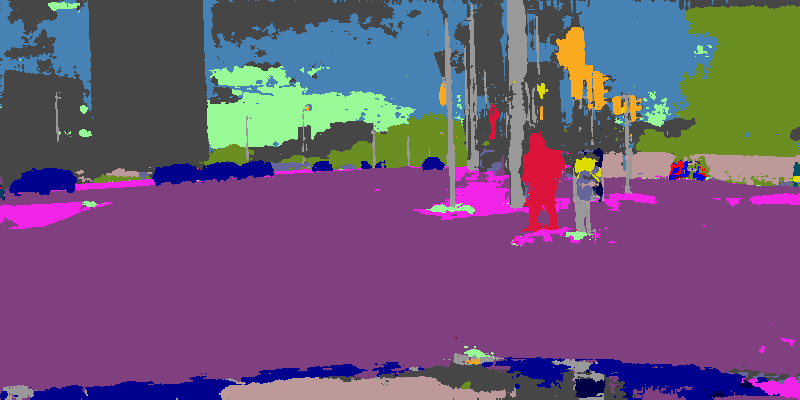}} & 
   \tablecenter{\includegraphics[width=\imgsize]{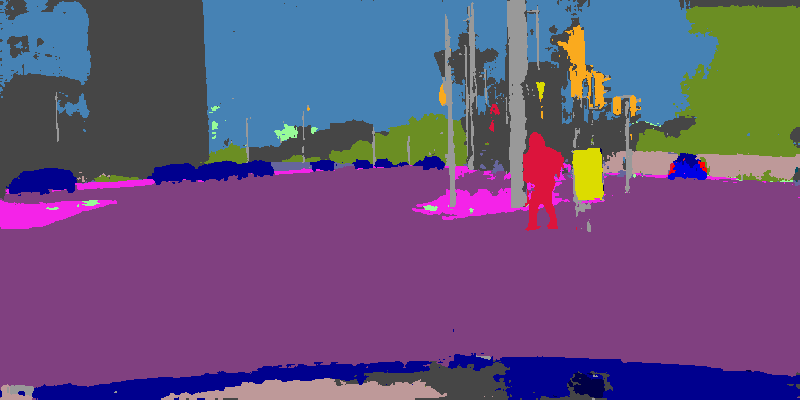}} \\
     & &   
   \tablecenter{\includegraphics[width=\imgsize]{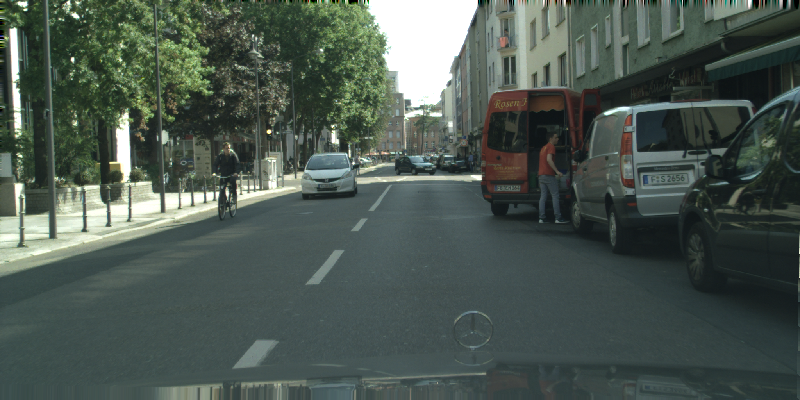}} &
   \tablecenter{\includegraphics[width=\imgsize]{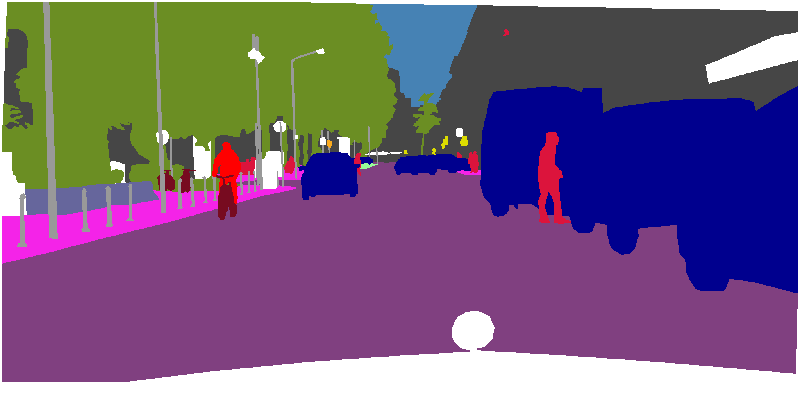}} & 
   \tablecenter{\includegraphics[width=\imgsize]{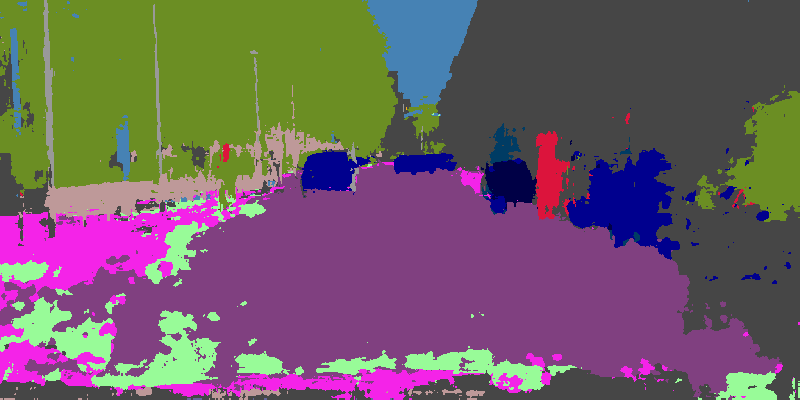}} & 
   \tablecenter{\includegraphics[width=\imgsize]{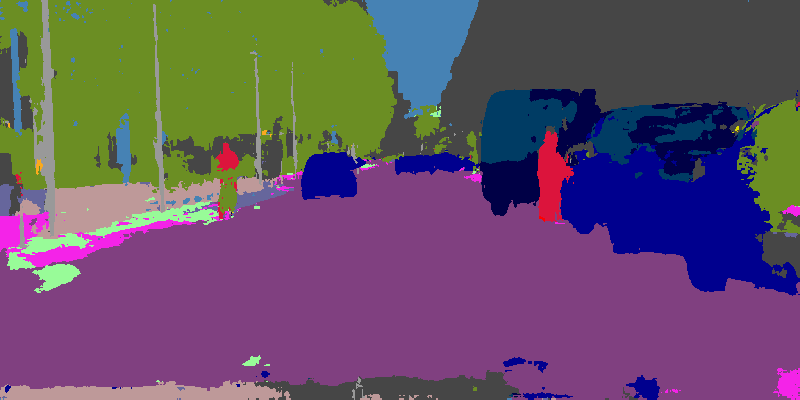}} & 
   \tablecenter{\includegraphics[width=\imgsize]{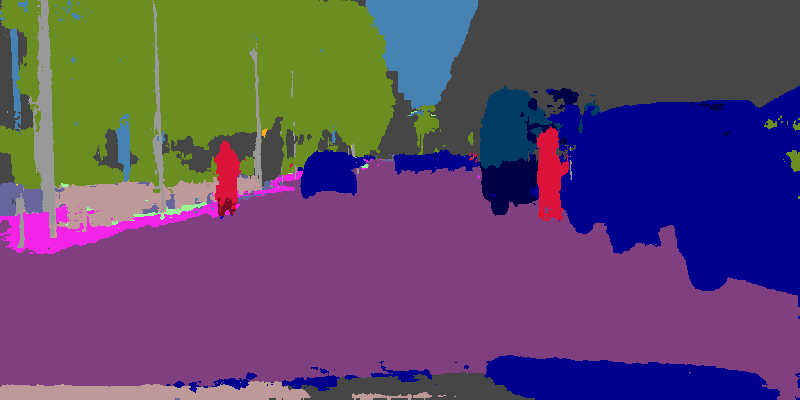}} \\
   
    & &   
   \tablecenter{\includegraphics[width=\imgsize]{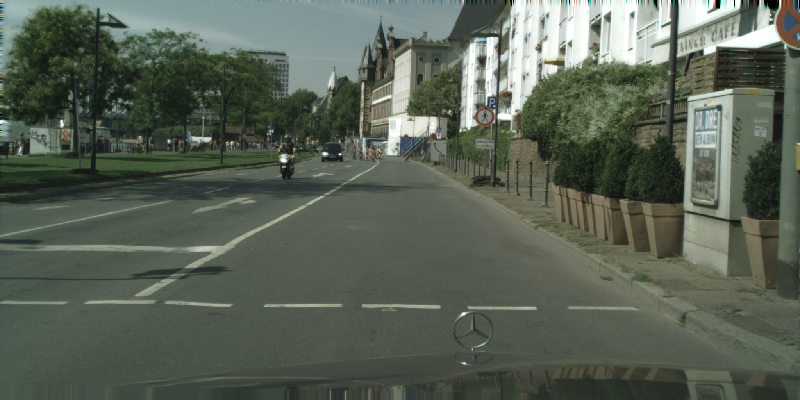}} &
   \tablecenter{\includegraphics[width=\imgsize]{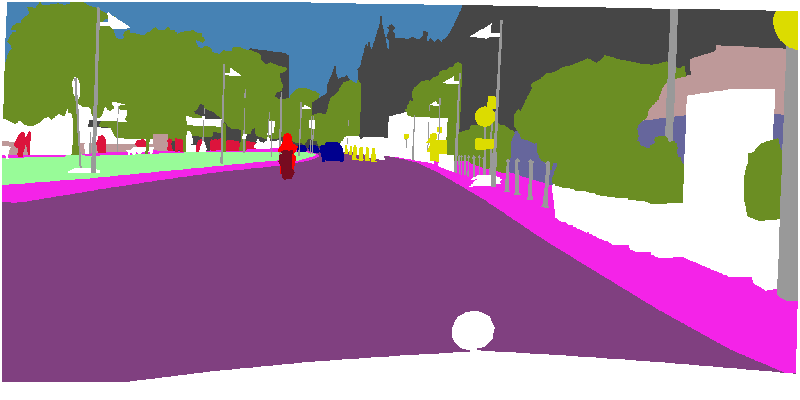}} & 
   \tablecenter{\includegraphics[width=\imgsize]{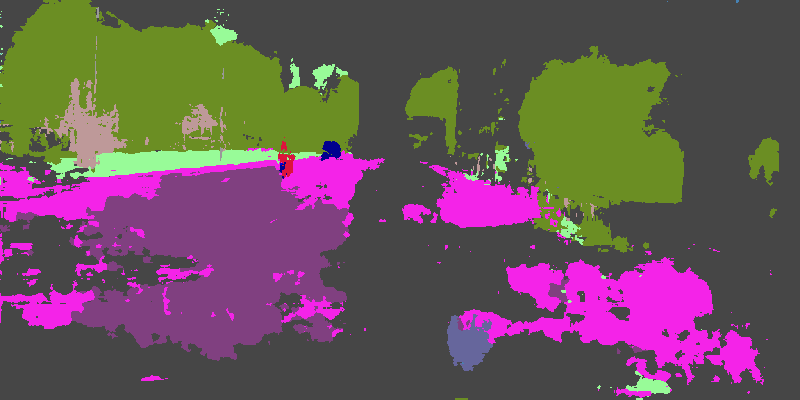}} & 
   \tablecenter{\includegraphics[width=\imgsize]{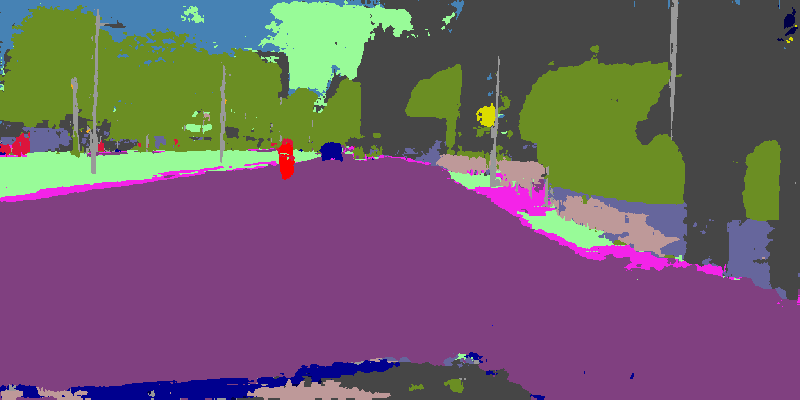}} & 
   \tablecenter{\includegraphics[width=\imgsize]{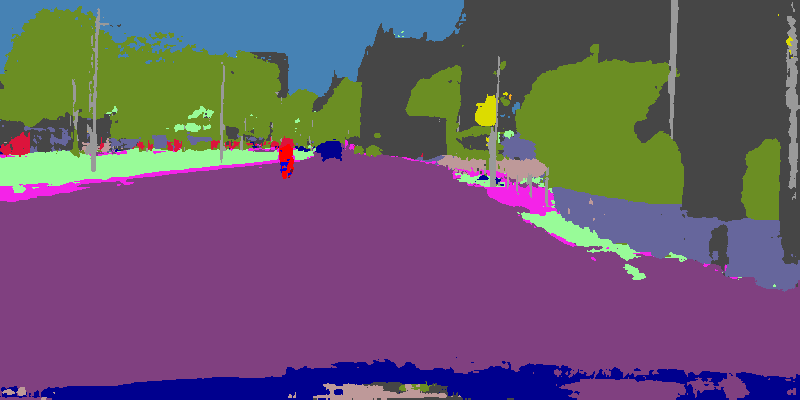}}\\ 
   \cline{2-2}
  
   & \multirow{4}{*}{\rotatebox{90}{\hspace{-15ex}From SYNTHIA}} & 
   \tablecenter{\includegraphics[width=\imgsize]{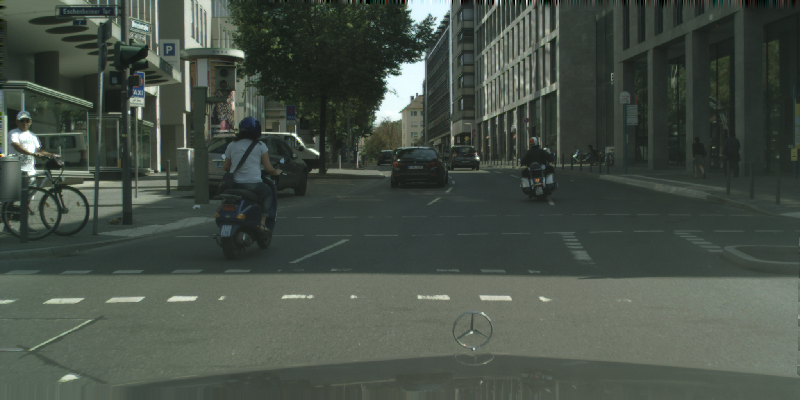}} &
   \tablecenter{\includegraphics[width=\imgsize]{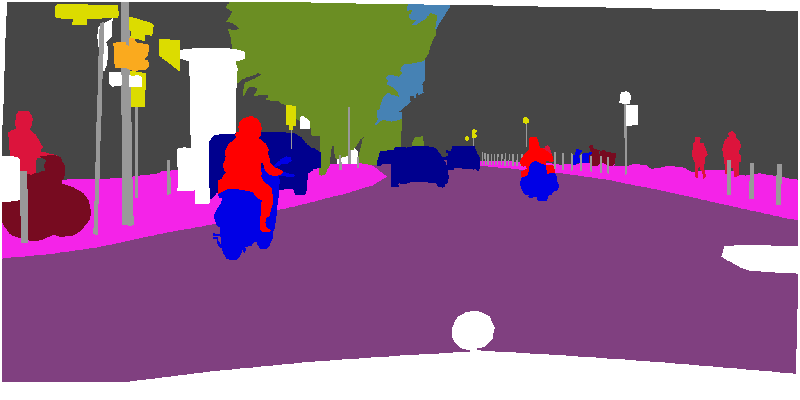}} & 
   \tablecenter{\includegraphics[width=\imgsize]{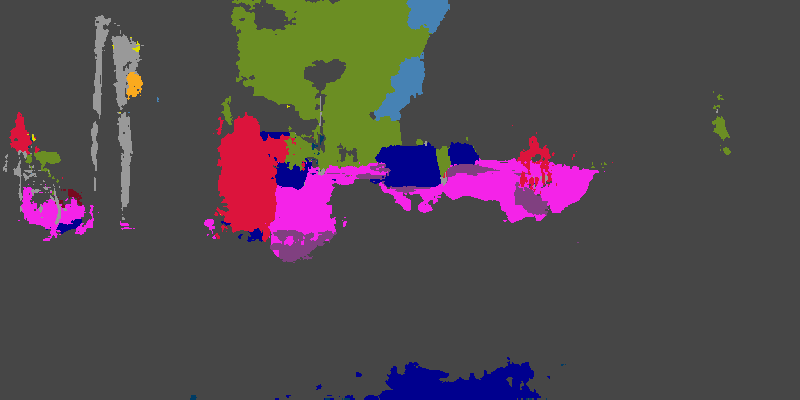}} & 
   \tablecenter{\includegraphics[width=\imgsize]{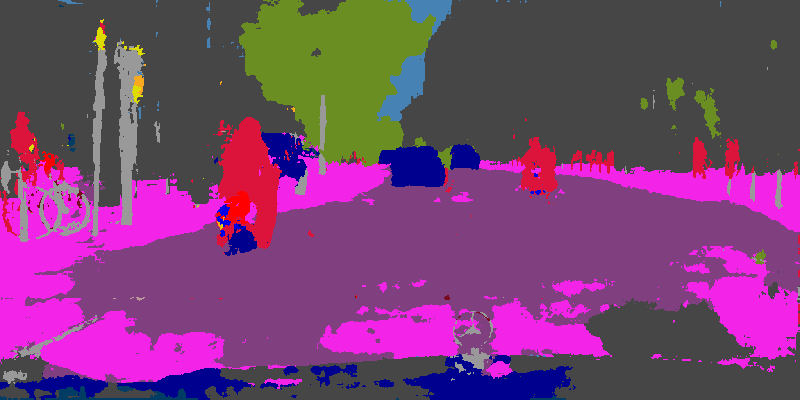}} & 
   \tablecenter{\includegraphics[width=\imgsize]{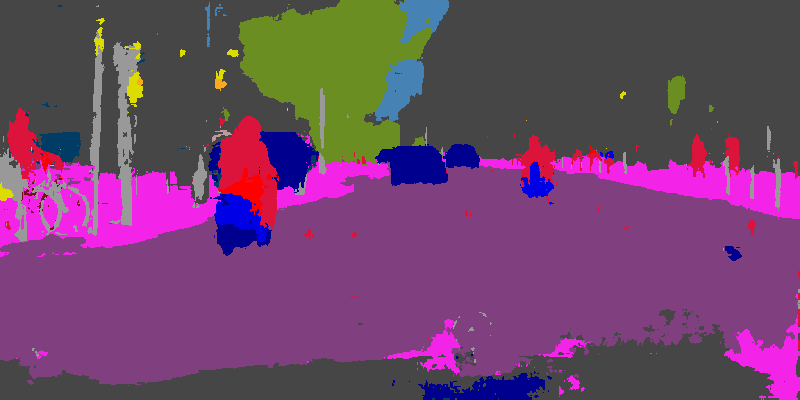}} \\
 
  & & 
   \tablecenter{\includegraphics[width=\imgsize]{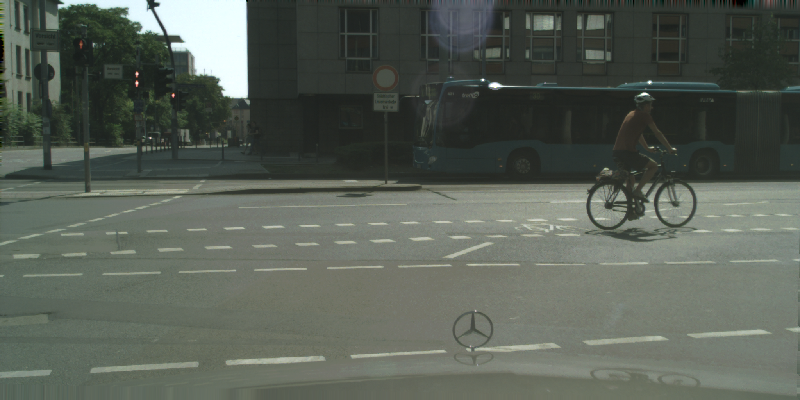}} &
   \tablecenter{\includegraphics[width=\imgsize]{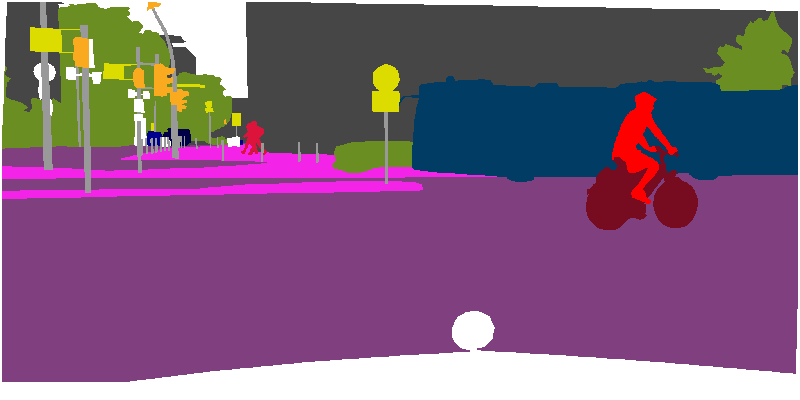}} & 
   \tablecenter{\includegraphics[width=\imgsize]{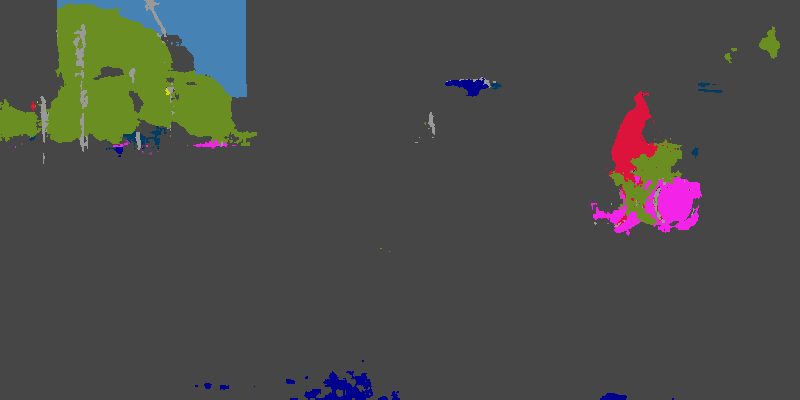}} & 
   \tablecenter{\includegraphics[width=\imgsize]{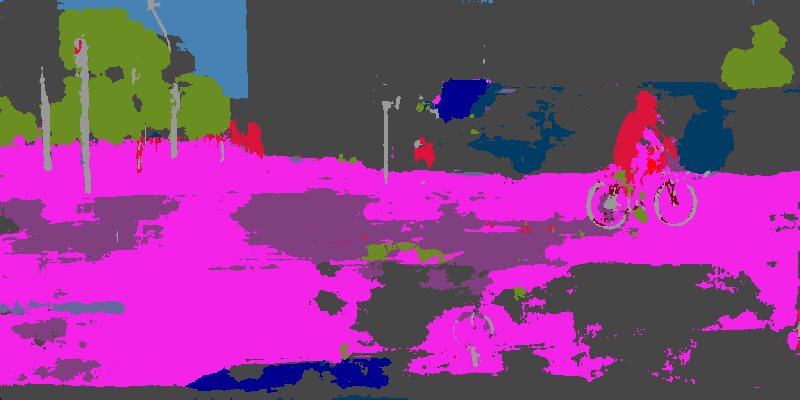}} & 
   \tablecenter{\includegraphics[width=\imgsize]{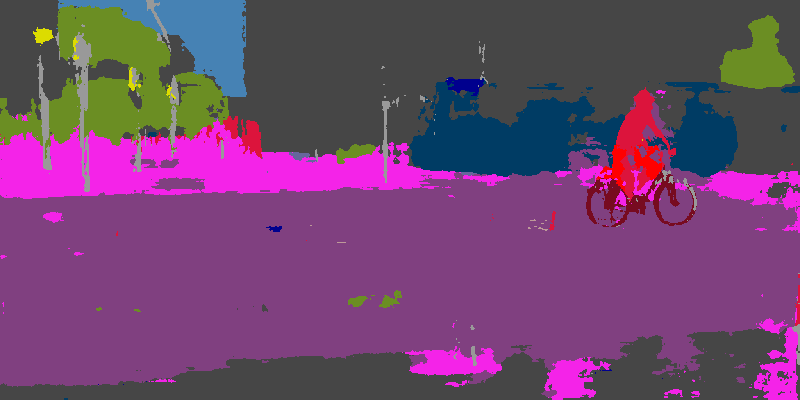}} \\
  
  & & 
   \tablecenter{\includegraphics[width=\imgsize]{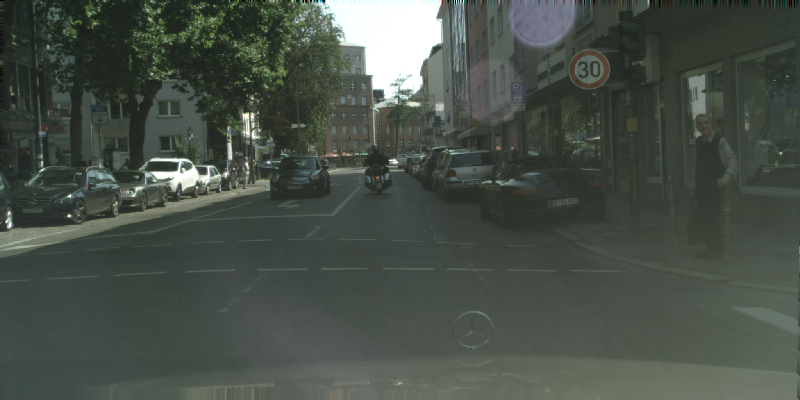}} &
   \tablecenter{\includegraphics[width=\imgsize]{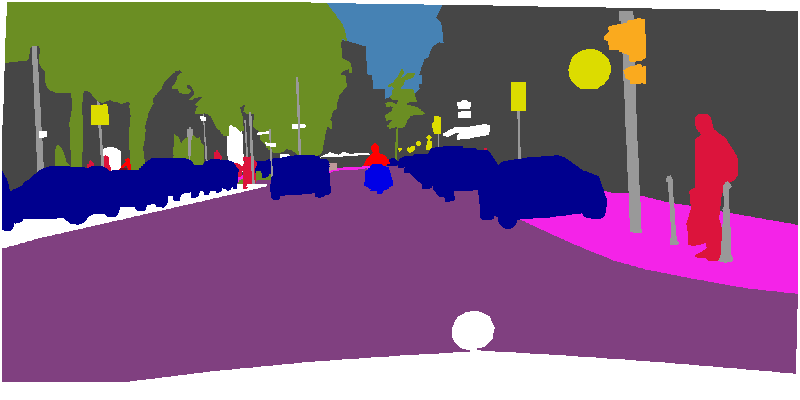}} & 
   \tablecenter{\includegraphics[width=\imgsize]{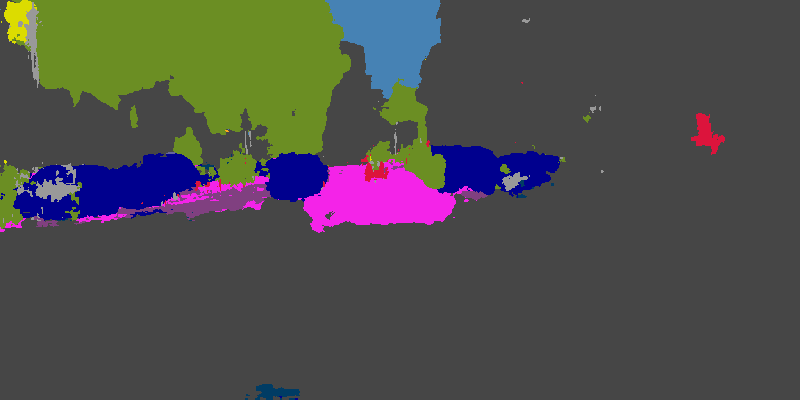}} & 
   \tablecenter{\includegraphics[width=\imgsize]{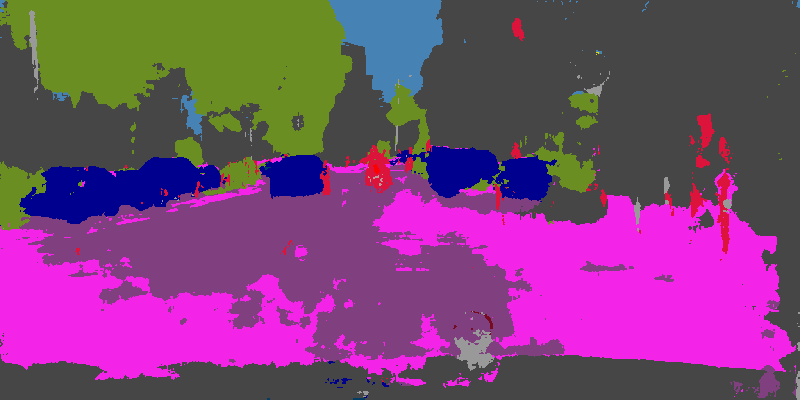}} & 
   \tablecenter{\includegraphics[width=\imgsize]{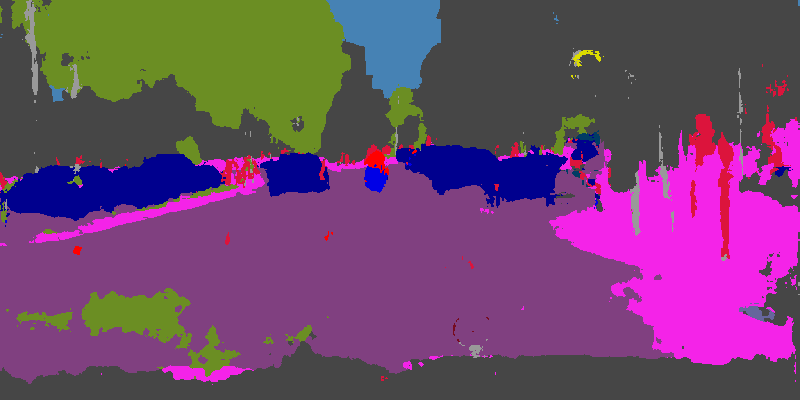}} \\
   
   & & 
   \tablecenter{\includegraphics[width=\imgsize]{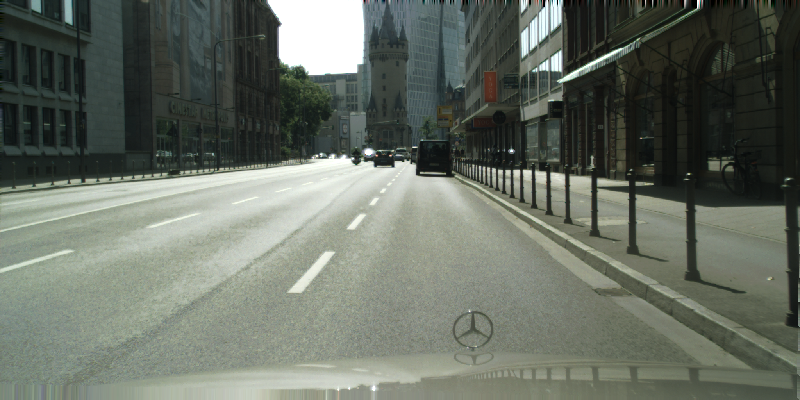}} &
   \tablecenter{\includegraphics[width=\imgsize]{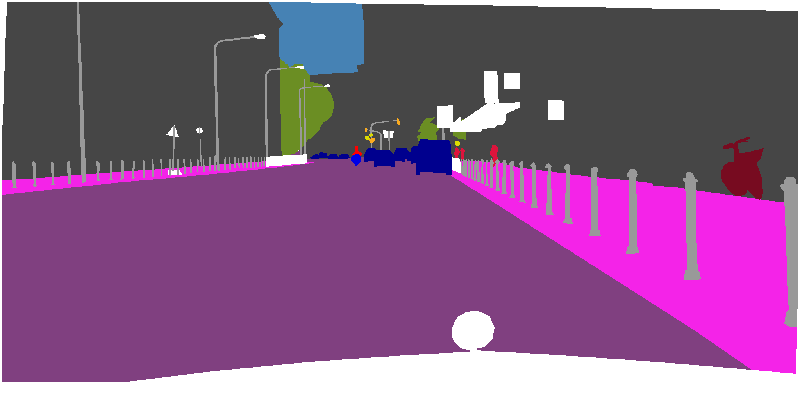}} & 
   \tablecenter{\includegraphics[width=\imgsize]{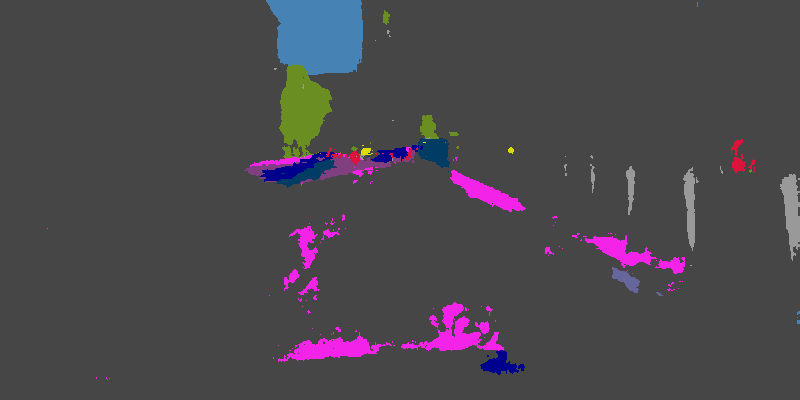}} & 
   \tablecenter{\includegraphics[width=\imgsize]{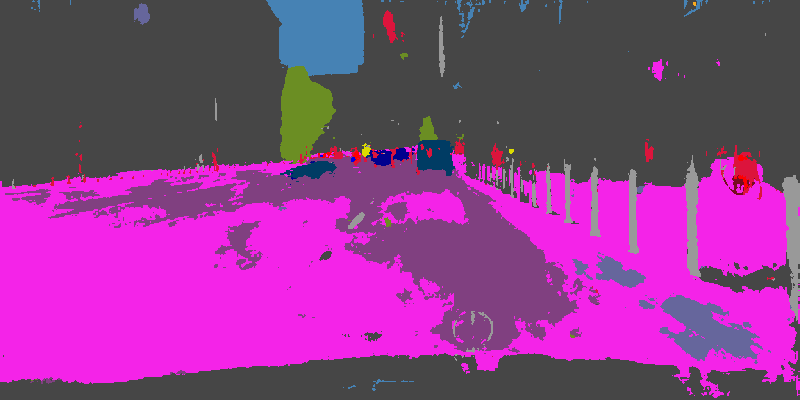}} & 
   \tablecenter{\includegraphics[width=\imgsize]{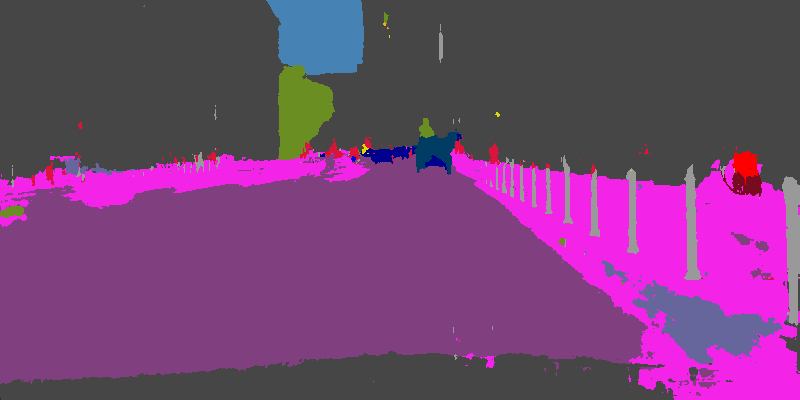}} \\
  \cline{1-2}
  \multicolumn{2}{c}{} & Image & Ground Truth & Source only & CyCADA \cite{hoffman2018} & Ours (full)
\end{tabular}
\end{subfigure}
\centering
\caption{Semantic segmentation of some sample scenes extracted from the Cityscapes validation set when adapting source knowledge learned on the GTA5 (rows $1-4$) and SYNTHIA (rows $5-8$) datasets (\textit{best viewed in colors}).}
\label{fig:qual_res}
\end{figure*}

\subsection{Ablation Study}

Finally, we analyze in detail the performance gain brought by the different components of our framework.  For this evaluation we considered the domain adaptation from GTA5 to Cityscapes.
A relevant contribution is due to the CycleGAN-based image-to-image translator, which effectively bridges the domain gap at the  pixel level by generating realistic target-like labeled data 
and pushes the accuracy on the Cityscapes validation set up to $39.1 \%$ (2nd row in Table \ref{tab:ablation}).  However, notice that this value is significantly different from the original CycleGAN \cite{cyclegan} result reported in Table \ref{tab:GTA} because in our framework the optimization of the segmentation network is made jointly with the cyclic translation and the training considers patches of images.
Then, we evaluated the impact  on the final adaptation performance  of the semantic and feature-based losses by alternately setting to $0$ the  $\lambda_{sem}$ and $\lambda_{feat}$ parameters.
The regularizing action of the feature level adaptation allows to increase the accuracy to $39.6 \%$  (3rd row in Table \ref{tab:ablation})  with a small but noticeable impact. 
The semantic loss (4th row in Table \ref{tab:ablation})  has a larger impact, leading to an improvement of $1.6 \%$ on the final score.
Finally, by using both components together we obtain a combined improvement of $2 \%$, leading to the final accuracy of $41.1 \%$.

\begin{table*}[htbp]
\setlength{\tabcolsep}{1.6pt}
\centering
\begin{tabular}{|c|ccccccccccccccccccc|c|}
\hline
 & \rotatebox{90}{road} &  \rotatebox{90}{sidewalk} &  \rotatebox{90}{building} &  \rotatebox{90}{wall} &\rotatebox{90}{fence} & \rotatebox{90}{pole} & \rotatebox{90}{t light} 
  &\rotatebox{90}{t sign} & \rotatebox{90}{veg} & \rotatebox{90}{terrain} & \rotatebox{90}{sky} & \rotatebox{90}{person}& \rotatebox{90}{rider} & \rotatebox{90}{car} 
  & \rotatebox{90}{truck} & \rotatebox{90}{bus} & \rotatebox{90}{train} &  \rotatebox{90}{mbike} & \rotatebox{90}{bike} & \rotatebox{90}{mean} \\
 \hline
Source only & 23.1 & 13.1 & 42.6 &  2.3 & 13.9 &  5.0 & 10.3 &  8.0 & 68.6 &  6.7 & 24.5 & 40.8 &  0.3 & 48.1 &  9.4 & 16.3 &  0.0 &  0.0 &  0.0 & 17.5 \\
Ours ($\lambda_{sem}, \lambda_{feat}=0$) & 87.8 & \textbf{39.1} & 81.0 & 24.8 & 16.0 & 31.9 & 27.2 & 25.7 & 78.6 & 22.9 & 69.7 & 55.5 & \textbf{16.7} & 85.3 & 25.7 & 31.0 &  \textbf{4.4} & 14.7 &  5.6 & 39.1 \\
Ours ($\lambda_{sem}=0$) & \textbf{88.3} & 37.1 & 81.1 & 25.4 & 15.3 & \textbf{33.6} & \textbf{29.5} & 28.8 & 80.0 & 24.4 & 69.2 & 56.0 & 15.6 & 85.0 & 25.5 & 30.8 &  3.9 & 16.3 &  6.2 & 39.6 \\
Ours ($\lambda_{feat}=0$) & 87.3 & 36.6 & 82.8 & \textbf{29.1} & \textbf{19.9} & 32.9 & 24.9 & 32.3 & 82.8 & 28.1 & 74.6 & 58.3 & 11.6 & \textbf{85.9} & \textbf{26.3} & \textbf{31.9} &  1.3 & \textbf{19.5} &  6.8 & 40.7 \\
Ours (full) & 87.6 & 36.7 & \textbf{83.5} & \textbf{29.1} & 17.8 & \textbf{33.6} & 24.3 & \textbf{35.2} & \textbf{83.1} & \textbf{28.9} & \textbf{76.3} & \textbf{59.1} & 14.0 & \textbf{85.9} & 25.4 & 29.4 &  2.6 & \textbf{19.5} &  \textbf{9.3} & \textbf{41.1} \\
\hline
Oracle      & 97.7 & 81.9 & 91.0 & 47.6 & 50.1 & 58.4 & 62.3 & 73.4 & 91.4 & 59.8 & 94.3 & 77.2 & 50.5 & 93.2 & 59.2 & 74.8 & 55.8 & 49.5 & 73.0 & 70.6\\
\hline
\end{tabular}
\caption{Ablation study on the Cityscapes validation set. The approaches have been trained in a supervised way on the GTA5 dataset and the unsupervised domain adaptation has been performed using the Cityscapes training set. The mean and per class highest results have been highlighted in bold.}
\label{tab:ablation}
\end{table*}

\section{Conclusions}
\label{sec:conclusions}

In this work we presented an effective method for unsupervised domain adaptation. We tested our approach on a challenging dense prediction task, i.e., adapting semantic segmentation from synthetic to real data in the autonomous driving scenario. 
Differently from other competing works, we built a unified framework combining cycle-consistency and adversarial domain adaptation both at image level and in the feature space. Our framework employs a MobileNet-v2 as segmentation network which is lightweight enough to allow a single shot end-to-end training on a single commercial GPU and to embed our full model in practical applications with limited computational constraints. We extensively validated our method on unsupervised adaptation starting from two widely used synthetic datasets. 

Future research will focus on the improvement of the feature level domain adaptation and to the inclusion of more advanced adversarial learning schemes for domain translation.

\section{Acknowledgment}
Our work was in part supported by the Italian Minister for Education (MIUR) under the  "Departments of Excellence" initiative (Law 232/2016) and by the University of Padova Strategic Research Infrastructure Grant 2017: “CAPRI: Calcolo ad Alte Prestazioni per la Ricerca e l’Innovazione”.

{
\bibliographystyle{elsarticle-num} 
\bibliography{strings,refs}
}

\end{document}